\documentclass[10pt,twocolumn,letterpaper]{article}

\usepackage[pagenumbers]{cvpr} 

\usepackage{graphicx}
\usepackage{booktabs}
\usepackage{amsmath}
\usepackage{amssymb}
\usepackage{times}
\usepackage{epsfig}
\usepackage{comment}
\usepackage{epigraph}
\usepackage{multirow}
\usepackage{hhline}
\usepackage{makecell}
\usepackage{bm}
\usepackage{subcaption}
\usepackage{soul}
\usepackage{array}
\usepackage{color,colortbl}

\makeatletter
\def\blfootnote{\gdef\@thefnmark{}\@footnotetext}
\makeatother

\newcommand{\argmin}{\operatornamewithlimits{argmin}}

\definecolor{beaublue}{rgb}{0.74, 0.83, 0.9}

\definecolor{cvprblue}{rgb}{0.21,0.49,0.74}
\usepackage[pagebackref,breaklinks,colorlinks,allcolors=cvprblue]{hyperref}

\def\confName{CVPR}
\def\confYear{2027}

\title{Unsupervised Skeleton-Based Action Segmentation via \\Hierarchical Spatiotemporal Vector Quantization}

\author{Umer Ahmed$^\dagger$~~~Syed Ahmed Mahmood$^\dagger$~~~Fawad Javed Fateh~~~M. Shaheer Luqman\\M. Zeeshan Zia~~~Quoc-Huy Tran\\
\\
Retrocausal, Inc.\\
Seattle, WA\\
\url{www.retrocausal.ai}
}

\begin{document}
\maketitle

\begin{abstract}
    We propose a novel hierarchical spatiotemporal vector quantization framework for unsupervised skeleton-based temporal action segmentation. We first introduce a hierarchical approach, which includes two consecutive levels of vector quantization. Specifically, the lower level associates skeletons with fine-grained subactions, while the higher level further aggregates subactions into action-level representations. Our hierarchical approach outperforms the non-hierarchical baseline, while primarily exploiting spatial cues by reconstructing input skeletons. Next, we extend our approach by leveraging both spatial and temporal information, yielding a hierarchical spatiotemporal vector quantization scheme. In particular, our hierarchical spatiotemporal approach performs multi-level clustering, while simultaneously recovering input skeletons and their corresponding timestamps. Lastly, extensive experiments on multiple benchmarks, including HuGaDB, LARa, and BABEL, demonstrate that our approach establishes a new state-of-the-art performance and reduces segment length bias in unsupervised skeleton-based temporal action segmentation.
\end{abstract}

\section{Introduction}
\label{sec:introduction}
{\blfootnote{$^{\dagger}$ indicates joint first author.\\ \{umer,ahmed,fawad,shaheer,zeeshan,huy\}@retrocausal.ai.}}

\begin{figure}[t]
	\centering
		\includegraphics[width=1.0\linewidth, trim = 0mm 45mm 10mm 0mm, clip]{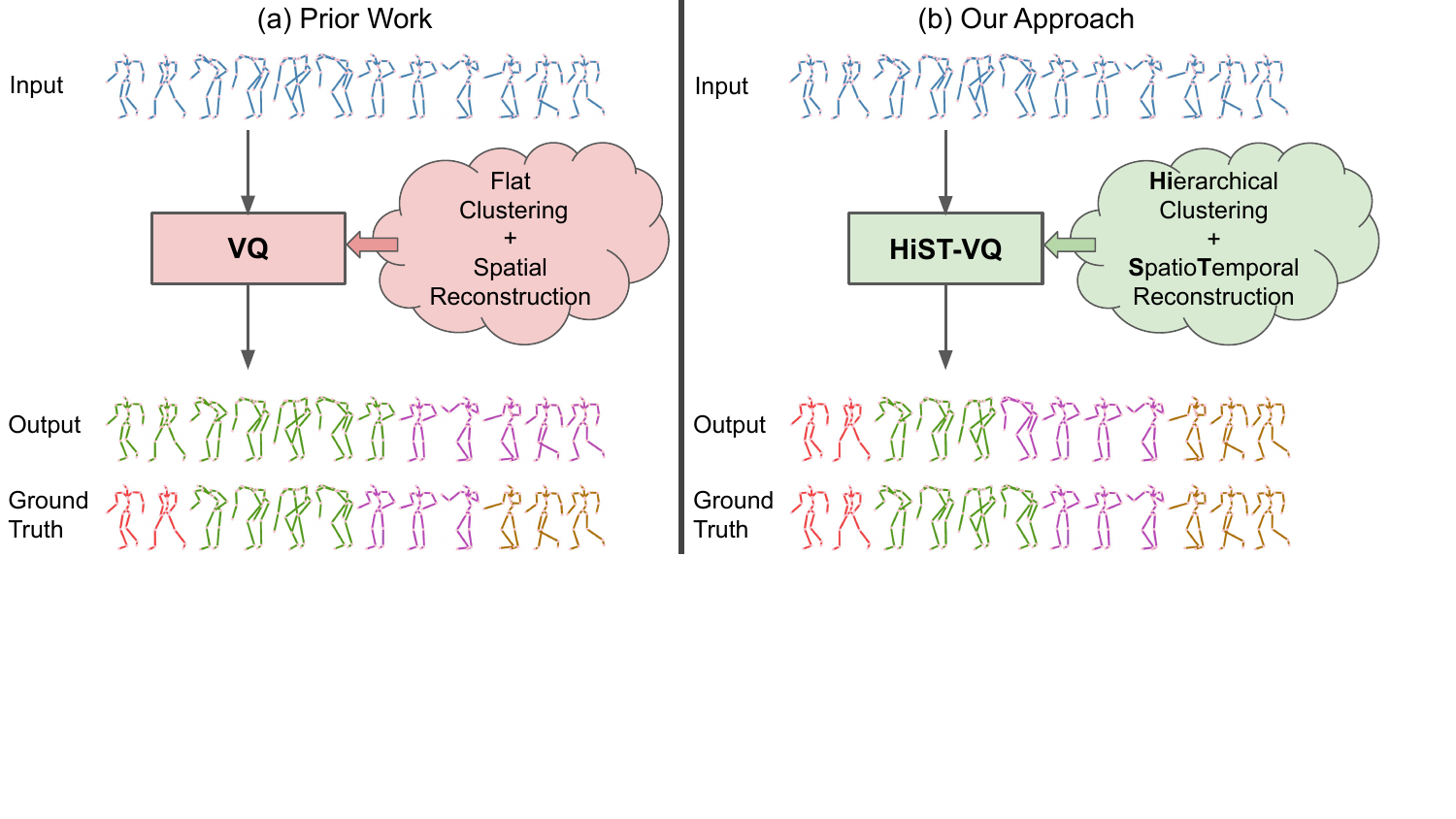}
	\caption{(a) Previous unsupervised skeleton-based temporal action segmentation methods, e.g., SMQ~\cite{gokay2025skeleton}, rely on traditional vector quantization techniques, which perform flat clustering and mostly exploit spatial cues via reconstructing input skeletons. (b) We propose a hierarchical spatiotemporal vector quantization approach, which conducts multi-level clustering and exploits both spatial and temporal cues by jointly reconstructing input skeletons and their associated timestamps. Our approach obtains better accuracy and less bias in predicted segment lengths. In the figures, output skeletons with the same color belong to the same action.}
	\label{fig:teaser}
\end{figure}

Interpreting human actions from skeletal motion data has become increasingly viable in recent years due to rapid advancements in motion capture and pose estimation techniques. Actions represented as 3D joint sequences provide a richer description of body structure and motion dynamics than RGB videos. Skeletons are also compact and largely free of appearance, background clutter, and camera viewpoint variations, allowing models to focus on body motion rather than nuisance factors~\cite{kwon2022context,tran2024learning}. Moreover, they are privacy-preserving as no raw images is processed, which is important for applications such as worker assistance and activity monitoring~\cite{kwon2022context,tran2024learning}. Segmenting skeleton sequences into discrete action units is also relevant to robotics, where decomposing human demonstrations supports learning from demonstration and action tokenization for robotic manipulation~\cite{chen2025moto, vuong2025action}. Despite these advantages, unsupervised skeleton-based temporal action segmentation remains challenging. Existing skeleton-based approaches are either fully supervised~\cite{filtjens2022skeleton, hosseini2020deep, ji2024language, li2023decoupled, tan2023hierarchical, tian2023stga, hyder2024action, tian2024spatial, xu2023efficient}, requiring costly frame-level annotations, or simplify the problem by segmenting short sequences that contain only a single action~\cite{guo2022contrastive, li20213d, lin2020ms2l, lin2023actionlet, paoletti2022unsupervised, su2020predict, zhang2022contrastive}. Recent works addressing unsupervised action segmentation of longer skeleton sequences containing many actions struggle to capture the intrinsic and temporal structure of complex actions.

Recent video-based approaches to unsupervised action segmentation have explored end-to-end clustering frameworks that jointly learn feature embeddings and cluster assignments~\cite{kumar2022unsupervised,tran2024permutation,xu2024temporally,bueno2025clot}. More recently, hierarchical vector quantization has proven effective in the RGB domain~\cite{spurio2025hierarchical}, highlighting the importance of capturing both fine-grained sub-actions and higher-level action representations that better reflect the compositional structure of human actions. Other video-based methods employ temporal reconstruction~\cite{kukleva2019unsupervised, vidalmata2021joint, li2021action}, yielding promising results and highlighting the importance of incorporating ordering information to produce temporally coherent segments. Some of these video-based techniques have been introduced in previous skeleton-based methods, e.g., SMQ~\cite{gokay2025skeleton}. However, they are limited to flat clustering and spatial reconstruction only (see Fig.~\ref{fig:teaser}(a)). Hierarchical vector quantization~\cite{spurio2025hierarchical} and temporal reconstruction~\cite{kukleva2019unsupervised, vidalmata2021joint, li2021action} remain largely unexplored in skeleton-based approaches. Moreover, directly applying video-based methods to skeleton data is suboptimal~\cite{gokay2025skeleton}, since skeletons are compact, structured joint sequences rather than dense appearance features, which motivates designs tailored to skeleton data.

We introduce HiST-VQ, a hierarchical spatiotemporal vector quantization approach for unsupervised skeleton-based temporal action segmentation. Instead of performing flat clustering over motion embeddings, HiST-VQ structures the clustering process across multiple levels, enabling it to discover short-term motion units that group together to form cohesive human actions. Furthermore, it jointly reconstructs the input skeletons together with timestamps, incorporating explicit temporal modeling, as shown in Fig.~\ref{fig:teaser}(b). This allows the learned representations to capture both structural pose information and temporal progression within long sequences. Lastly, as we demonstrate in Sec.~\ref{sec:comparisons}, our model achieves state-of-the-art performance and reduces segment length bias (i.e., prior works tend to predict long segments and fail to capture short segments --- Fig.~\ref{fig:teaser}(a) shows an example where the first and last segments are overlooked) than previous methods in unsupervised skeleton-based temporal action segmentation. 

In summary, our contributions include:
\begin{itemize}
    \item Firstly, we develop a hierarchical approach for unsupervised skeleton-based temporal action segmentation based on hierarchical vector quantization. Our hierarchical approach outperforms the non-hierarchical counterpart, while focusing on spatial information through reconstructing input skeletons.
    \item Secondly, we further exploit temporal cues by jointly recovering input skeletons and their timestamps, yielding a hierarchical spatiotemporal approach.
    \item Finally, extensive evaluations on several datasets show that our hierarchical spatiotemporal approach achieves higher accuracy and less bias in predicted segment lengths compared to prior works.
\end{itemize}
\section{Related Work}
\label{sec:relatedwork}

\noindent \textbf{Video-Based Action Segmentation.}
Notable research efforts have been invested in video-based action segmentation. Supervised methods~\cite{lea2017temporal,ding2017tricornet,lei2018temporal,farha2019ms,khan2022timestamp} often use Temporal Convolutional Networks (TCNs) and require framewise/weak labels for full/weak supervision. Unsupervised methods have been proposed to mitigate labeling challenges. Early attempts~\cite{sener2015unsupervised,alayrac2016unsupervised} exploit narrations accompanying videos. However, narrations are not always available. To address that, visual-based methods~\cite{sener2018unsupervised,kukleva2019unsupervised,vidalmata2021joint,li2021action,kumar2022unsupervised,tran2024permutation,xu2024temporally,spurio2025hierarchical,bueno2025clot,ali2025joint} have been developed. CTE~\cite{kukleva2019unsupervised} learns a temporal embedding and employs K-means to cluster embedded features. VTE~\cite{vidalmata2021joint} and ASAL~\cite{li2021action} add visual embeddings and action embeddings to improve CTE. These methods separate representation learning and offline clustering. Recently, TOT~\cite{kumar2022unsupervised} proposes a joint representation learning and online clustering framework. ASOT~\cite{xu2024temporally} relaxes balanced assignment constraints in TOT. USFA~\cite{tran2024permutation} and CLOT~\cite{bueno2025clot} extend TOT~\cite{kumar2022unsupervised} and ASOT~\cite{xu2024temporally} by utilizing segment cues. More recently, HVQ~\cite{spurio2025hierarchical} notes a segment length bias in these methods and introduces a hierarchical vector quantization approach to alleviate that. Despite promising results on video data, applying the above methods directly to skeleton data does not yield optimal results~\cite{gokay2025skeleton}. Based on insights from video-based methods~\cite{kukleva2019unsupervised,spurio2025hierarchical}, we propose a skeleton-based approach which obtains state-of-the-art performance with less segment length bias.

\noindent \textbf{Skeleton-Based Action Segmentation.}
Skeleton-based action segmentation has attracted increasing research attention~\cite{yan2018spatial,parsa2020spatio,parsa2021multi,filtjens2022skeleton,liu2022spatial,tian2023stga,xu2023efficient,li2023decoupled,li2023involving,ji2024language,gokay2025skeleton}, thanks to the compactness and privacy-preserving nature of skeleton data. Several works~\cite{yan2018spatial,parsa2020spatio,parsa2021multi,filtjens2022skeleton} integrate Graph Convolutional Networks (GCNs) with Temporal Convolutional Networks (TCNs) to jointly model spatial and temporal dynamics. Recently, a few methods~\cite{li2023decoupled,li2023involving} which explicitly decouple spatial and temporal representations have been developed, while various attention mechanisms~\cite{liu2022spatial,tian2023stga} have been introduced to better capture spatiotemporal dependencies. Other approaches focus on alternative paradigms, such as action synthesis~\cite{xu2023efficient} and the incorporation of language priors~\cite{ji2024language}. All of the aforementioned methods require labels for supervised training. More recently, SMQ~\cite{gokay2025skeleton} presents an unsupervised skeleton-based action segmentation approach based on a classical vector quantization formulation, which involves only flat clustering and spatial reconstruction. Here, we propose a hierarchical spatiotemporal vector quantization framework, which performs hierarchical clustering and spatiotemporal reconstruction, yielding higher accuracy and reduced segment length bias.

\noindent \textbf{Self-Supervised Learning with Skeleton Data.}
Self-supervised representation learning~\cite{chen2020simple, he2020momentum, grill2020bootstrap, caron2018deep, caron2020unsupervised, caron2021emerging, he2022masked, feichtenhofer2022masked} leverages pretext tasks that exploit the inherent structure of unlabeled data to learn meaningful feature embeddings. This strategy has been extended to skeleton-based action recognition~\cite{su2020predict, xu2021unsupervised, zhang2022contrastive, guo2022contrastive, lin2023actionlet, li20213d, zheng2018unsupervised, lin2020ms2l}, which typically focuses on short clips depicting a single action. For example, CrosSCLR~\cite{li20213d}, AimCLR~\cite{guo2022contrastive}, and ActCLR~\cite{lin2023actionlet} employ contrastive learning frameworks to learn discriminative representations. These methods emphasize feature extraction alone and do not incorporate action label prediction tasks during training. Consequently, they still require annotated data for downstream tasks and are restricted to short, single-action sequences. Recently, LAC~\cite{yang2023lac} builds on pre-trained visual encoders to model compositional actions using synthesized data, while hBehaveMAE~\cite{stoffl2024elucidating} employs a hierarchical masked autoencoding architecture to capture interpretable latent action representations across different granularities. Both models are fine-tuned with labeled data before evaluation on downstream tasks. Unlike these methods, our approach identifies and segments actions directly during training in a fully unsupervised setting.

\noindent \textbf{Vector Quantized Variational Autoencoders.}
Vector Quantized Variational Autoencoders (VQ-VAE)~\cite{van2017neural} learn discrete latent representations by mapping continuous encoder outputs to a finite codebook of embedding vectors and jointly training with reconstruction and commitment losses. This enables interpretable discrete codes for complex data while mitigating posterior collapse in traditional VAEs. This framework has been applied to several computer vision tasks, including image generation~\cite{razavi2019generating}, video generation~\cite{yan2021videogpt}, action recognition~\cite{chen2025masked}, and action segmentation~\cite{spurio2025hierarchical,gokay2025skeleton}. These methods mostly use single-level codebooks and primarily exploit spatial cues via spatial reconstruction losses. In this work, we propose a hierarchical spatiotemporal framework that employs multi-level codebooks and leverages both spatial and temporal cues via spatiotemporal reconstruction losses, establishing a new state-of-the-art and reducing segment length bias in unsupervised skeleton-based action segmentation. Our method is also strongly connected to action tokenization for robot manipulation~\cite{chen2025moto, vuong2025action}, where discrete codes serve as reusable action primitives. Similarly, our hierarchical codebooks yield discrete subaction and action units from unlabeled skeleton sequences, which could be used for such tokenization. We will explore this in more detail in our future work.

\section{Our Approach}
\label{sec:method}

\begin{figure*}[t]
	\centering
		\includegraphics[width=0.65\linewidth, trim = 0mm 0mm 0mm 0mm, clip]{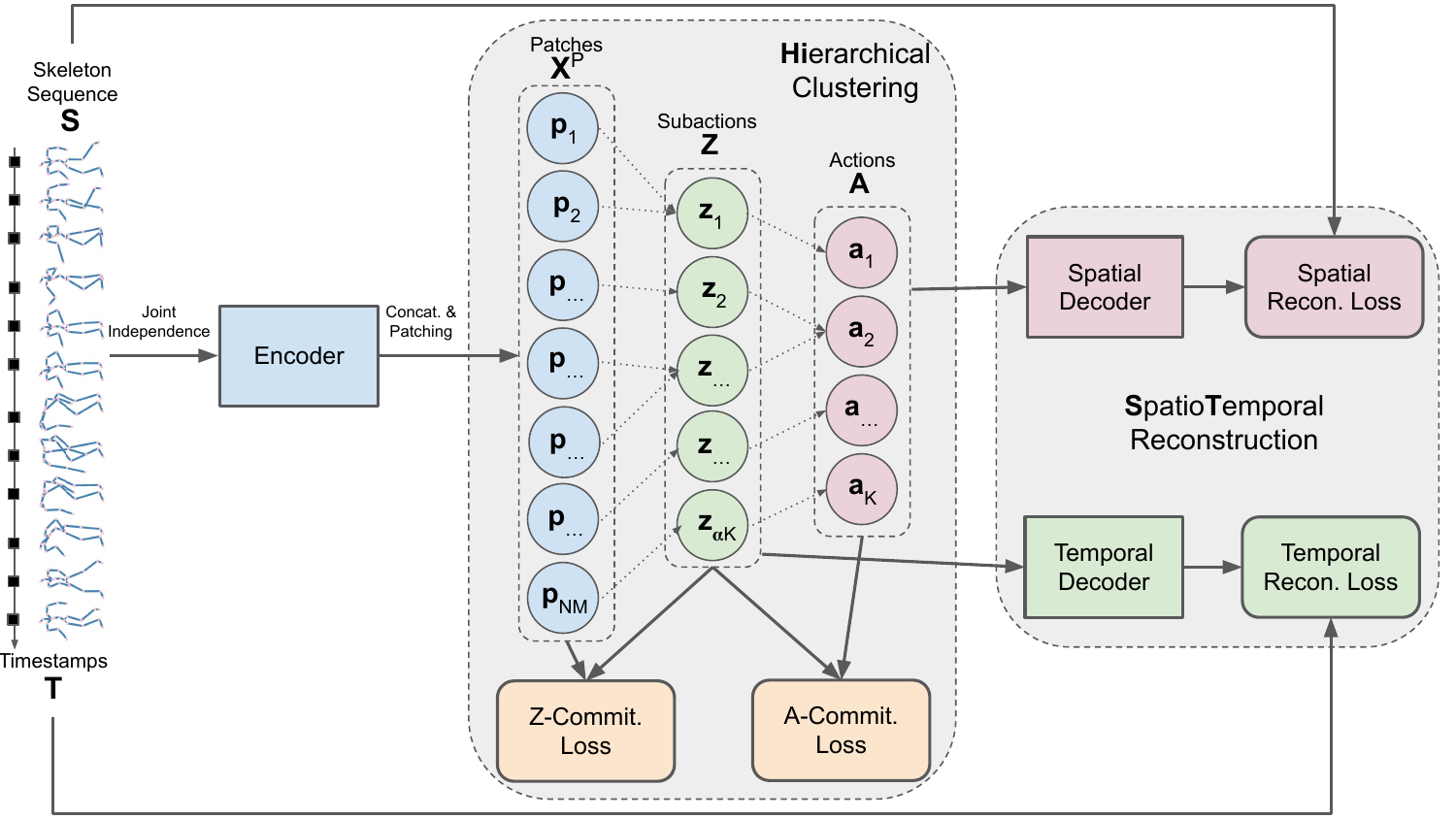}
	\caption{Given an input skeleton sequence $\mathbf{S}$ with associated timestamps $\mathbf{T}$, we pass $\mathbf{S}$ to an encoder, which maps each joint sequence independently to the latent space. We concatenate the embedded joint sequences into the embedded skeleton sequence and divide it into patches along the temporal dimension, yielding $\mathbf{X}^P$. Next, each patch $\mathbf{p}_k$ is first assigned to the nearest subaction prototype $\mathbf{z}_j$, which is then assigned to the closest action prototype $\mathbf{a}_i$. Lastly, quantized action patches $\mathbf{Q}_A$ are passed to a spatial decoder to reconstruct the input skeleton sequence, while quantized subaction patches $\mathbf{Q}_Z$ are fed to a temporal decoder to recover the input timestamps. The encoder, decoders, and prototypes are learned jointly via a combination of spatial reconstruction, temporal reconstruction, and commitment losses.}
	\label{fig:method}
\end{figure*}

Unsupervised temporal action segmentation aims to divide unlabeled videos into temporally coherent segments and clusters them into semantically meaningful actions within and across videos. Considerable research efforts have been devoted to developing unsupervised temporal action segmentation methods for video data, whereas unsupervised skeleton-based approaches have only recently emerged, due to the robustness and privacy-preserving advantage of skeleton data. Motivated by unsupervised video-based methods~\cite{kukleva2019unsupervised,spurio2025hierarchical}, we present our main contribution in this section, HiST-VQ, a \textbf{Hi}erarchical \textbf{S}patio\textbf{T}emporal \textbf{V}ector \textbf{Q}uantization framework for unsupervised skeleton-based temporal action segmentation. HiST-VQ is not a direct transfer of video-based techniques, which perform poorly on skeleton data (Sec.~\ref{sec:comparisons}). We build the hierarchy on patch-based skeleton representations, and the two levels serve distinct roles, with action-level codes used for spatial reconstruction and subaction-level codes used for temporal reconstruction (Sec.~\ref{sec:ablations}). Our approach consists of two key modules: i) hierarchical clustering, which first maps skeletons to subactions and then maps subactions to actions, and ii) spatiotemporal reconstruction, which learns self-supervised representations via reconstructing both input skeletons and corresponding timestamps. By leveraging these modules, our approach not only obtains state-of-the-art performance but also reduces segment length bias. Fig.~\ref{fig:method} illustrates an overview of HiST-VQ. We provide our model details and training losses in Secs.~\ref{sec:model_details} and ~\ref{sec:training_losses} respectively.

\subsection{Model Details}
\label{sec:model_details}

\noindent \textbf{Patch-Based Representation.}
We define input skeleton sequences as $\mathbf{S} \in \mathbb{R}^{N \times C \times T \times V}$, where $N$, $C$, $T$, and $V$ respectively denote the number of sequences (or batch size), joint dimension (e.g., $C = 3$ for 3D joints), sequence length, and number of joints. We follow SMQ~\cite{gokay2025skeleton} to learn an embedding for each joint independently. Particularly, $\mathbf{S}$ is reshaped into $\mathbf{S}' \in \mathbb{R}^{N \times V \times C \times T}$. Each joint sequence $\mathbf{S}'_{nv} \in \mathbb{R}^{C \times T}$ is processed separately by an encoder to capture joint-specific motion patterns, yielding the embedded joint sequence $\mathbf{X}_{nv} \in \mathbb{R}^{D \times T}$, with $D$ representing the latent dimension. The encoder is a Multi-Stage Temporal Convolutional Network (TCN)~\cite{farha2019ms} composed of dilated residual layers that progressively refine temporal features. Each stage applies a $1 \times 1$ convolution for feature projection, followed by residual blocks with exponentially increasing dilation to capture multi-scale temporal dependencies. A final $1 \times 1$ convolution projects the features to the latent space. Moreover, we transform skeleton-wise representations to patch-wise representations to capture temporal variability more effectively, following~\cite{gokay2025skeleton}. Specifically, we first aggregate the above embedded joint sequences $\mathbf{X}_{nv}$ into the embedded skeleton sequences $\mathbf{X}' \in \mathbb{R}^{N \times T \times V \times D}$, and then split $\mathbf{X}'$ into non-overlapping patches along the temporal dimension $T$, yielding the patch sequences $\mathbf{X}^P \in \mathbb{R}^{N \times M \times P \times V \times D}$, where $P$ is the patch size and $M = T/P$ is the number of patches in each sequence.

\noindent \textbf{Hierarchical Clustering.}
Inspired by HVQ~\cite{spurio2025hierarchical}, we present a \emph{patch-based} hierarchical vector quantization framework. Our vector quantization hierarchy consists of two learned patch-based codebooks $\mathbf{Z} = \{\mathbf{z}_j\}^{\alpha K}_{j=1}$ and $\mathbf{A} = \{\mathbf{a}_i\}^{K}_{i=1}$, corresponding to two levels of vector quantization. Here, $\mathbf{z}_j \in \mathbb{R}^{P \times V \times D}$, $\mathbf{a}_i \in \mathbb{R}^{P \times V \times D}$, $K$ is the number of actions, and $\alpha$ is a ratio parameter. $\mathbf{A}$ represents $K$ action prototypes/clusters, while $\mathbf{Z}$ models $\alpha K$ subaction prototypes/clusters. The first vector quantization level maps each patch $\mathbf{p}_k \in \mathbf{X}^P$ to the closest prototype $\mathbf{z}_{j^*} \in \mathbf{Z}$, yielding the quantized $\mathbf{q}^Z_k$ as:
\begin{align}
    \mathbf{q}^Z_k = \mathbf{z}_{j^*},~~~\text{with}~~~j^* = \argmin_j ||\mathbf{p}_k - \mathbf{z}_j||_2.
\end{align}
Merging $\mathbf{q}^Z_k$ from all $\mathbf{p}_k \in \mathbf{X}^P$ and then depatchifying yield the quantized $\mathbf{Q}^Z \in \mathbb{R}^{N \times T \times V \times D}$.
Similarly, the second vector quantization level maps the prototype $\mathbf{q}^Z_k \in \mathbf{Z}$ to the nearest prototype $\mathbf{a}_{i^*} \in \mathbf{A}$, yielding the quantized $\mathbf{q}^A_k$ as:
\begin{align}
    \mathbf{q}^A_k = \mathbf{a}_{i^*},~~~\text{with}~~~i^* = \argmin_i ||\mathbf{q}^Z_k - \mathbf{a}_i||_2.
\end{align}
Combining $\mathbf{q}^A_k$ from all $\mathbf{p}_k \in \mathbf{X}^P$ and then depatchifying produce the quantized $\mathbf{Q}^A \in \mathbb{R}^{N \times T \times V \times D}$. Following~\cite{van2017neural}, we apply Exponential Moving Average (EMA) to update the learned patch-based codebooks as:

\begin{equation}
\begin{aligned}
    \mathbf{\hat{z}}_j = \frac{1}{\hat{N}_{\mathbf{z}_j}} \left( \beta \mathbf{z}_j + (1-\beta) \sum_{\mathbf{q}^Z_k = \mathbf{z}_j} \mathbf{p}_k \right),\\~~~\hat{N}_{\mathbf{z}_j} = \beta N_{\mathbf{z}_j} + (1-\beta)|\{ \mathbf{q}^Z_k = \mathbf{z}_j \}|,
\end{aligned}
\end{equation}

\begin{equation}
\begin{aligned}
    \mathbf{\hat{a}}_i = \frac{1}{\hat{N}_{\mathbf{a}_i}} \left( \beta \mathbf{a}_i + (1-\beta) \sum_{\mathbf{q}^A_k = \mathbf{a}_i} \mathbf{q}^Z_k \right),\\~~~\hat{N}_{\mathbf{a}_i} = \beta N_{\mathbf{a}_i} + (1-\beta)|\{ \mathbf{q}^A_k = \mathbf{a}_i \}|.
\end{aligned}
\end{equation}

Here, $N_{\mathbf{z}_j}$ is the prior estimate of $\hat{N}_{\mathbf{z}_j}$, corresponding to the number of assigned patches to prototype $\mathbf{z}_j$. Similarly, $N_{\mathbf{a}_i}$ is the previous estimate of $\hat{N}_{\mathbf{a}_i}$, representing the number of assigned prototypes to prototype $\mathbf{a}_i$. When $\hat{N}_{\mathbf{z}_j} < \nu_z$ and $\hat{N}_{\mathbf{a}_i} < \nu_a$ for several batches, we replace $\mathbf{z}_j$ and $\mathbf{a}_i$ with random inputs sampled within the current batch~\cite{dhariwal2020jukebox}. As discussed in Sec.~\ref{sec:comparisons}, our hierarchical approach achieves superior performance with less segment length bias than the non-hierarchical baseline of SMQ~\cite{gokay2025skeleton}.

\noindent \textbf{Spatiotemporal Reconstruction.}
Motivated by CTE~\cite{kukleva2019unsupervised}, we introduce \emph{patch-based} spatiotemporal reconstruction as our pretext task for self-supervised learning, which exploits both spatial and temporal cues through simultaneously recovering the input skeleton sequences and associated timestamps. In particular, for spatial reconstruction, we first rearrange the quantized action patches $\mathbf{Q}^A$ for joint independence before passing them to a spatial decoder, which follows the encoder's architecture in reverse, producing the reconstructed skeletons $\mathbf{\hat{S}} \in \mathbb{R}^{N \times C \times T \times V}$ after reshaping. Furthermore, for temporal reconstruction, we first reshape the quantized subaction patches $\mathbf{Q}^Z$ and then feed them to a temporal decoder, which has a simpler architecture (i.e., an MLP network with 2 hidden layers) than the spatial decoder, yielding the predicted timestamps $\mathbf{\hat{T}} \in \mathbb{R}^{N \times M}$. Note that we predict a timestamp for each patch, instead of each frame. As studied in Sec.~\ref{sec:ablations}, utilizing the quantized action patches $\mathbf{Q}^A$ for spatial reconstruction yields the best results, while using the quantized subaction patches $\mathbf{Q}^Z$ for temporal reconstruction performs the best. By leveraging both spatial and temporal cues, our approach with spatiotemporal reconstruction outperforms the spatial reconstruction baseline of SMQ~\cite{gokay2025skeleton}, as demonstrated in Sec.~\ref{sec:comparisons}. 

\noindent \textbf{Differences with Prior Work.}
There are several differences between our method and prior work, including: 1) Compared to SMQ~\cite{gokay2025skeleton}, which performs flat clustering and spatial reconstruction only, HiST-VQ performs two-level hierarchical clustering and jointly recovers skeletons and timestamps. 2) Compared to HVQ~\cite{spurio2025hierarchical}, which is developed in the RGB domain, HiST-VQ builds its hierarchical codebooks on patch-based skeleton representations. 3) Compared to CTE~\cite{kukleva2019unsupervised}, which separates representation learning and offline clustering, HiST-VQ learns the encoder, decoders, and both codebooks jointly in an end-to-end manner. 4) Our decoders are tied to different levels of the hierarchy, i.e., spatial reconstruction from action-level codes and temporal reconstruction from subaction-level codes, a design absent from the above methods and validated in Sec.~\ref{sec:ablations}. Each component contributes to our results, as shown by the ablations on hierarchy levels, temporal reconstruction, and decoder inputs in Sec.~\ref{sec:ablations}.

\subsection{Training Losses}
\label{sec:training_losses}

We train our encoder, spatial decoder, temporal decoder, subaction codebook, and action codebook using a combination of hierarchical clustering and spatiotemporal reconstruction losses. The codebooks are randomly initialized.

\noindent \textbf{Hierarchical Clustering.}
For hierarchical clustering, we employ two \emph{patch-based} commitment losses, corresponding to the two vector quantization levels in Fig.~\ref{fig:method}, as:
\begin{align}
    \mathcal{L}_{commit_Z} = \sum^{N \cdot M}_{k=1} || \mathbf{p}_k - \text{sg}[\mathbf{q}^Z_k]||^2_2,
\end{align}
\begin{align}
    \mathcal{L}_{commit_A} = \sum^{N \cdot M}_{k=1} || \mathbf{q}^Z_k - \text{sg}[\mathbf{q}^A_k]||^2_2.
\end{align}
$\mathcal{L}_{commit_Z}$ encourages patch $\mathbf{p}_k$ to stay close to the assigned prototype $\mathbf{q}^Z_k$, while $\mathcal{L}_{commit_A}$ pushes prototype $\mathbf{q}^Z_k$ towards the chosen prototype $\mathbf{q}^A_k$. sg[$\cdot$] denotes stop-gradient, and $N \cdot M$ is the total number of patches in $\mathbf{X}^P$.

\noindent \textbf{Spatiotemporal Reconstruction.}
To measure spatial reconstruction errors between reconstructed skeletons $\mathbf{\hat{S}} \in \mathbb{R}^{N \times C \times T \times V}$ and original skeletons $\mathbf{S} \in \mathbb{R}^{N \times C \times T \times V}$, we adopt the inter-joint distance Mean Squared Error (MSE) loss~\cite{gokay2025skeleton}, which is written as:


\begin{equation}
\begin{aligned}
    \mathcal{L}_{spat} = \frac{1}{N \cdot T \cdot V^2} \sum^N_{n=1} \sum^T_{t=1} \sum^V_{v=1} &\sum^V_{w=1} ( || \mathbf{S}_{ntv} - \mathbf{S}_{ntw}||^2_2 \\
    &- || \mathbf{\hat{S}}_{ntv} - \mathbf{\hat{S}}_{ntw}||^2_2 )^2.
\end{aligned}
\end{equation}

Our \emph{patch-based} temporal reconstruction loss is defined as MSE between predicted timestamps $\mathbf{\hat{T}}  \in \mathbb{R}^{N \times M}$ and original timestamps $\mathbf{T} \in \mathbb{R}^{N \times M}$ and is expressed as:
\begin{align}
    \mathcal{L}_{temp} = \frac{1}{N \cdot M} \sum^N_{n=1} \sum^M_{m=1} (\mathbf{T}_{nm} - \mathbf{\hat{T}}_{nm})^2.
\end{align}
The above simple-yet-effective temporal loss and decoder improve the results in Sec.~\ref{sec:experiments}. Complex alternatives may further improve the results, which remains our future work.

\noindent \textbf{Final Loss.}
Our final loss merges all of the above losses:
\begin{equation}
\begin{aligned}
    \mathcal{L} = \lambda_{commit}(\mathcal{L}_{commit_Z} + \mathcal{L}_{commit_A}) &+ \lambda_{spat}\mathcal{L}_{spat} \\
    &+ \lambda_{temp}\mathcal{L}_{temp}.
\end{aligned}
\end{equation}

Here, $\lambda_{commit}$ is the weight for the hierarchical clustering losses, while 
$\lambda_{spat}$ and $\lambda_{temp}$ are the weights for the spatial and temporal reconstruction losses respectively.
\section{Experiments}
\label{sec:experiments}

\noindent \textbf{Datasets.}
We evaluate our HiST-VQ model on 3 publicly available skeleton-based action segmentation datasets. Namely, HuGaDB~\cite{chereshnev2017hugadb}, LARa~\cite{niemann2020lara} and BABEL~\cite{punnakkal2021babel}.
\begin{itemize}
    \item \textbf{HuGaDB} is a human gait analysis dataset consisting of 10 hours of recordings of 10 lower limb actions, e.g., walking, running, walking up, and down stairs. The data were collected from 18 participants wearing a body sensor network consisting of 6 3-axis inertial sensors.
    \item \textbf{LARa} is a sensor-based human action recognition dataset for logistics optimization containing 8 action classes. It comprises 13 hours of recordings of 14 subjects, using full-body MoCap to track the position and orientation of 22 joints in 3D space. The skeleton is normalized by centering it at the root joint for translation invariance. The entire dataset is downsampled from 200 FPS to 50 FPS.
    \item \textbf{BABEL} labels 43 hours of MoCap sequences from the AMASS~\cite{mahmood2019amass} dataset. It consists of 3D skeleton motion sequences with 63K framewise labels of 250 action classes. We divide the dataset into 3 subsets, focusing on 4 action classes each (totaling 12 classes) and 25 full-body joints. This is in accordance with the setup in~\cite{yu2023frame}. Similarly to LARa, we also downsample this dataset to 30 FPS and center the skeletons at the root joint.
\end{itemize}

\noindent \textbf{Implementation Details.}
We employ two-stage TCN~\cite{farha2019ms} with 3 dilated residual layers for each stage as our encoder and spatial decoder, and MLP with 2 hidden layers as our temporal decoder. We adopt a two-level vector quantization hierarchy and set $\alpha = 2$, making the codebook size twice the number of ground truth classes ($K$) for the first level and the same as the number of ground truth classes for the second level. An exponential moving average (EMA)~\cite{van2017neural} is used for codebook updates with a decay weight $\beta$ = 0.5. We set the codebook usage thresholds $\nu_z = 3$ and $\nu_a = 1$. We feed $\mathbf{Q}^A$ to the spatial decoder, and $\mathbf{Q}^Z$ to the temporal decoder. We set $\lambda_{commit} = 1$ and $\lambda_{spat} = 0.001$ for all datasets. The patch size is 60 frames for HuGaDB, 50 frames for LARa, and 30 frames for BABEL, corresponding to 1 second for each dataset. We train our model using ADAM optimizer~\cite{kingma2014adam} with a learning rate of $5\times10^{-4}$. The training is run on a single NVIDIA RTX 3090 Ti (24GB).

\noindent \textbf{Competing Methods.}
We compare against state-of-the-art unsupervised skeleton-based temporal action segmentation model SMQ~\cite{gokay2025skeleton}, and unsupervised video-based models with skeleton data input CTE~\cite{kukleva2019unsupervised}, TOT~\cite{kumar2022unsupervised}, ASOT~\cite{xu2024temporally}, and HVQ~\cite{spurio2025hierarchical}. For CTE and TOT, we also compare their enhanced versions with Viterbi decoding~\cite{kuehne2018hybrid, ding2023temporal}, selecting the most probable action sequence under a first-order Markov model and penalizes frequent label changes.

\noindent \textbf{Evaluation Metrics.}
Following prior works~\cite{gokay2025skeleton, kukleva2019unsupervised, kumar2022unsupervised, xu2024temporally,spurio2025hierarchical}, predicted segments are matched to ground truth action labels using the global Hungarian matching algorithm applied over the entire dataset, after which the evaluation metrics are computed.
We report Mean over Frames (MoF), which measures the percentage of correctly predicted frames. However, MoF does not penalize over-segmentation and is biased toward longer and more frequent actions, making it less sensitive to errors on short segments. Therefore, we additionally report the Edit score, as well as the segmental F1-score evaluated at IoU thresholds of 10\%, 25\%, and 50\%, which provide a more comprehensive evaluation while penalizing over-segmentation.

\subsection{Comparison Results}
\label{sec:comparisons}

\noindent \textbf{Results on HuGaDB.}
The quantitative results on HuGaDB are summarized in \cref{tab:Results_on_Hugadb_Lara}. Our proposed model HiST-VQ achieves the best performance across all metrics among all unsupervised approaches, establishing a new state of the art. In particular, HiST-VQ achieves a MoF of 48.2 and an Edit score of 44.3, outperforming the previous best method SMQ~\cite{gokay2025skeleton} by 6.2 and 8.2 points respectively. Consistent improvements are also observed across all F1 thresholds, with gains of 10.9, 8.3, and 4.0 points at F1@10, F1@25, and F1@50 as compared to SMQ~\cite{gokay2025skeleton}. Although supervised methods still have higher performance, they require costly framewise labels. In contrast, our method achieves good performance without any supervision, highlighting its effectiveness for scalable temporal action segmentation where annotations are unavailable.

\begin{table}[tb]
    \caption{Results on HuGaDB and LARa, including supervised and unsupervised approaches.}
    \label{tab:Results_on_Hugadb_Lara}
    \centering
    \setlength{\tabcolsep}{3pt}
    \resizebox{\columnwidth}{!}{
    \begin{tabular}{@{}l|lllll|lllll@{}}
        \hline
        & \multicolumn{5}{c|}{\textbf{HuGaDB}} & \multicolumn{5}{c}{\textbf{LARa}} \\
        \cline{2-11}
        \textbf{Method} & \textbf{MoF} & \textbf{Edit} & \multicolumn{3}{c|}{\textbf{F1@\{10, 25, 50\}}} & \textbf{MoF} & \textbf{Edit} & \multicolumn{3}{c}{\textbf{F1@\{10, 25, 50\}}} \\
        \hline
        \textbf{Supervised} & & & & & & & & & &\\
        \hspace{1em}TCN \cite{lea2017temporal} & 88.3 & - & - & - & 56.8 & 61.5 & - & - & - & 20.0 \\
        \hspace{1em}ST-GCN \cite{yan2018spatial} & 88.7 & - & - & - & 67.7 & 67.9 & - & - & - & 25.8 \\
        \hspace{1em}MS-TCN \cite{farha2019ms} & 86.8 & - & - & - & 89.9 & 65.8 & - & - & - & 39.6 \\
        \hspace{1em}MS-GCN \cite{filtjens2022skeleton} & 90.4 & - & - & - & 93.0 & 65.6 & - & - & - & 43.6 \\
        \hline
        \textbf{Unsupervised} & & & & & & & & & &\\
        \hspace{1em}CTE \cite{kukleva2019unsupervised} & 33.8 & 4.7 & 0.6 & 0.6 & 0.5 & 23.3 & 16.8 & 8.1 & 5.2 & 2.3 \\
        \hspace{1em}CTE+Viterbi \cite{kukleva2019unsupervised} & 39.2 & 21.7 & 13.2 & 9.5 & 7.5 & 23.0 & 17.7 & 6.8 & 3.7 & 1.6 \\
        \hspace{1em}TOT \cite{kumar2022unsupervised} & 33.8 & 3.1 & 0.7 & 0.5 & 0.4 & 21.4 & 7.8 & 5.3 & 2.7 & 1.1 \\
        \hspace{1em}TOT+Viterbi \cite{kumar2022unsupervised}  & 33.8 & 20.8 & 15.6 & 10.5 & 7.5 & 32.6 & 17.7 & 11.6 & 7.4 & 3.2 \\
        \hspace{1em}ASOT \cite{xu2024temporally} & 33.9 & 17.4 & 4.5 & 3.8 & 3.0 & 22.9 & 23.4 & 17.8 & 12.1 & 5.7 \\
        \hspace{1em}HVQ \cite{spurio2025hierarchical} & 26.0 & 24.8 & 13.4 & 6.3 & 2.2 & 33.2 & 17.0 & 11.0 & 4.1 & 1.1 \\
        \hspace{1em}SMQ \cite{gokay2025skeleton} & \underline{42.0} & \underline{36.1} & \underline{38.5} & \underline{31.5} & \underline{24.3} & \underline{37.4} & \underline{39.4} & \underline{34.7} & \underline{28.4} & \underline{16.4} \\
        \cellcolor{beaublue}\hspace{1em}\textbf{HiST-VQ (Ours)} & \cellcolor{beaublue}\textbf{48.2} & \cellcolor{beaublue}\textbf{44.3} & \cellcolor{beaublue}\textbf{49.4} & \cellcolor{beaublue}\textbf{39.8} & \cellcolor{beaublue}\textbf{28.3} & \cellcolor{beaublue}\textbf{45.9} & \cellcolor{beaublue}\textbf{42.1} & \cellcolor{beaublue}\textbf{41.1} & \cellcolor{beaublue}\textbf{34.0} & \cellcolor{beaublue}\textbf{19.3} \\
        \hline
    \end{tabular}
    }
\end{table}

\noindent \textbf{Results on LARa.}
\cref{tab:Results_on_Hugadb_Lara} also presents the results on LARa. It is clear from \cref{tab:Results_on_Hugadb_Lara} that HiST-VQ performs the best across all metrics, yielding a new state of the art. Our method obtains a MoF of 45.9 and an Edit score of 42.1, improving over the closest competitor  SMQ~\cite{gokay2025skeleton}, by 8.5 and 2.7 points respectively. Across the F1 metrics, our approach consistently outperforms SMQ~\cite{gokay2025skeleton}, showing increases of 6.4 at F1@10, 5.6 at F1@25, and 2.9 at F1@50. These results further demonstrate the advantage of our hierarchical spatiotemporal vector quantization over the baseline method SMQ~\cite{gokay2025skeleton}.

\noindent \textbf{Results on BABEL.}
We now report the performance on BABEL in \cref{tab:Results_on_Babel}. While our approach outperforms previous methods on MoF and F1 across all subsets, our Edit is worse than HVQ~\cite{spurio2025hierarchical} on the BABEL subsets 2 and 3. Nevertheless, our model achieves the best overall performance across all subsets. Furthermore, it is evident from \cref{tab:Results_on_Babel} that HiST-VQ consistently outperforms the baseline method SMQ~\cite{gokay2025skeleton} across all metrics and subsets, which validates the benefit of our multi-level clustering and spatiotemporal reconstruction modules.

\begin{table*}[tb]
    \caption{Results on BABEL subsets.}
    \label{tab:Results_on_Babel}
    \centering

    \scriptsize
    \renewcommand{\arraystretch}{1.1}
    \setlength{\tabcolsep}{3pt}
    
    \begin{tabular}{@{}l|lllll|lllll|lllll@{}}
        \hline
        & \multicolumn{5}{c|}{\textbf{BABEL Subset-1}} & \multicolumn{5}{c|}{\textbf{BABEL Subset-2}} & \multicolumn{5}{c}{\textbf{BABEL Subset-3}} \\
        \cline{2-16}
        \textbf{Method} & \textbf{MoF} & \textbf{Edit} & \multicolumn{3}{c|}{\textbf{F1@\{10, 25, 50\}}} & \textbf{MoF} & \textbf{Edit} & \multicolumn{3}{c|}{\textbf{F1@\{10, 25, 50\}}} & \textbf{MoF} & \textbf{Edit} & \multicolumn{3}{c}{\textbf{F1@\{10, 25, 50\}}} \\
        \hline
        CTE \cite{kukleva2019unsupervised} & 34.8 & 28.6 & 25.0 & 17.5 & 9.5 &
        40.3 & 30.6 & 17.8 & 12.2 & 7.4 &
        31.4 & 13.1 & 8.2 & 5.8 & 3.6 \\
        
        CTE+Viterbi \cite{kukleva2019unsupervised} & 30.9 & 36.2 & 23.2 & 15.2 & 7.3 
        & 42.4 & 30.7 & 24.3 & 19.5 & 12.8
        & 31.2 & 30.9 & 20.7 & 15.2 & 8.4 \\
        
        TOT \cite{kumar2022unsupervised} & 31.8 & 18.7 & 14.2 & 7.6 & 4.4 & 
        35.4 & 12.8 & 13.7 & 8.6 & 4.3 & 
        31.5 & 7.1 & 4.9 & 2.9 & 1.7 \\
        
        TOT+Viterbi \cite{kumar2022unsupervised}  & 29.1 & 29.3 & 31.5 & 20.8 & 9.9 &
        35.3 & 36.8 & 35.9 & 30.0 & 19.8 &
        34.0 & 33.8 & 31.3 & 26.8 & 17.9\\
        
        ASOT \cite{xu2024temporally} & 35.3 & \underline{43.1} & \underline{42.3} & \underline{34.1} & \underline{24.5} &
        43.1 & 37.7 & 40.3 & 33.4 & 23.4 &
        38.0 & 27.1 & 27.4 & 21.6 & 14.3 \\

        HVQ \cite{spurio2025hierarchical} & \underline{43.2} & 15.0 & 10.6 & 5.3 & 1.7 &
        \underline{51.3} & \textbf{53.3} & \underline{49.3} & \underline{40.1} & 26.4 & 
        37.6 & \textbf{51.5} & \underline{46.8} & \underline{33.1} & 14.6 \\
        
        SMQ \cite{gokay2025skeleton} & 36.6 & 38.5 & 40.9 & 32.8 & 22.3 &
        49.1 & 37.8 & 43.8 & 37.4 & \underline{27.4} &
        \underline{40.6} & 38.6 & 38.0 & 29.3 & \underline{19.3} \\

        \cellcolor{beaublue}\textbf{HiST-VQ (Ours)} & \cellcolor{beaublue}\textbf{44.1} & \cellcolor{beaublue}\textbf{43.4} & \cellcolor{beaublue}\textbf{44.5} & \cellcolor{beaublue}\textbf{36.0} & \cellcolor{beaublue}\textbf{24.7} &
        \cellcolor{beaublue}\textbf{58.3} & \cellcolor{beaublue}\underline{43.6} & \cellcolor{beaublue}\textbf{51.8} & \cellcolor{beaublue}\textbf{41.1} & \cellcolor{beaublue}\textbf{30.4} &
        \cellcolor{beaublue}\textbf{44.0} & \cellcolor{beaublue}\underline{41.8} & \cellcolor{beaublue}\textbf{47.1} & \cellcolor{beaublue}\textbf{33.2} & \cellcolor{beaublue}\textbf{24.1} \\
        \hline
    \end{tabular}
\end{table*}

\begin{figure*}[t]
	\centering
		\includegraphics[width=0.85\linewidth, trim = 0mm 80mm 0mm 0mm, clip]{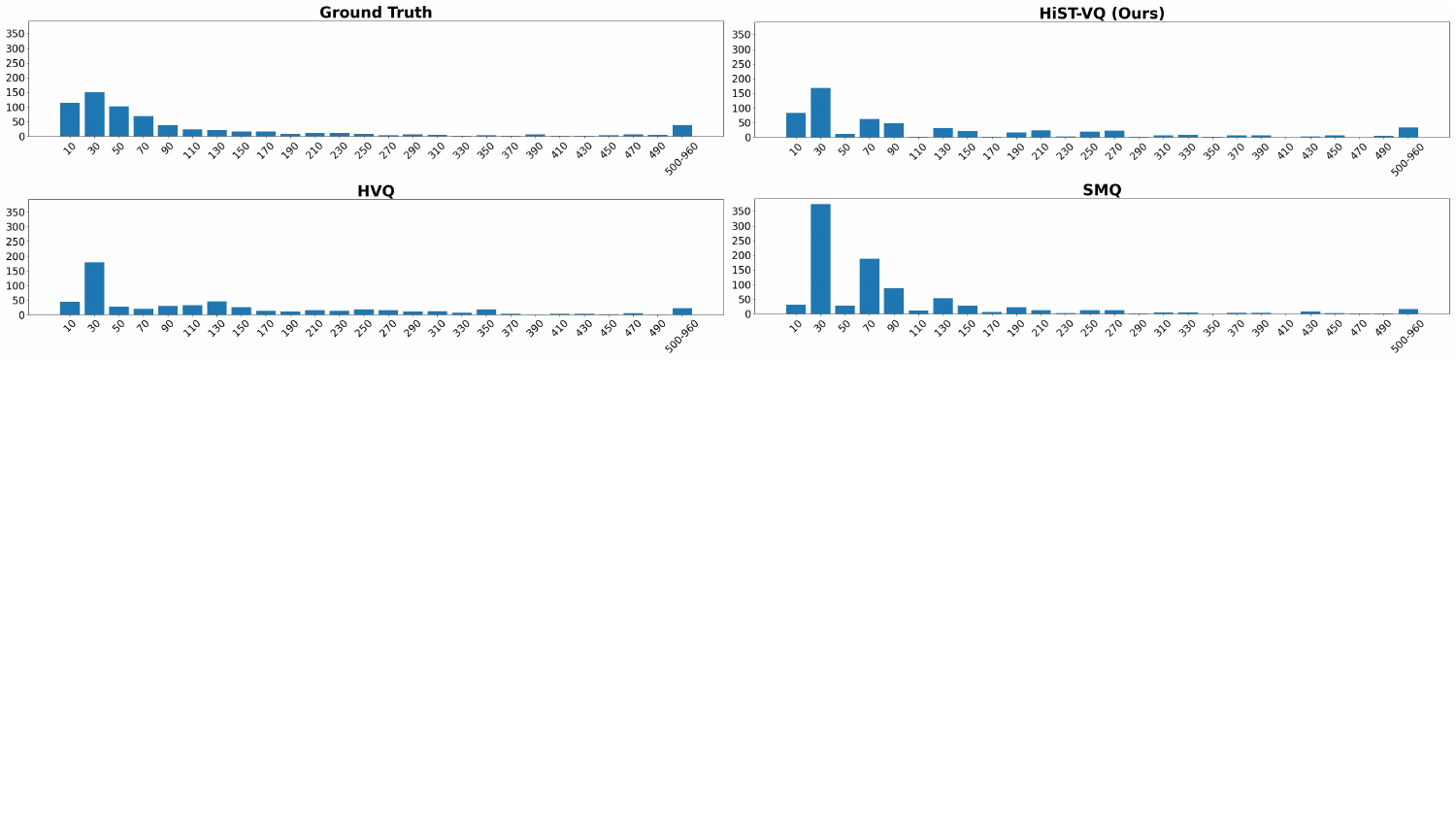}
	\caption{Histograms of segment lengths on BABEL Subset-3.}
	\label{fig:histogram}
\end{figure*}

\noindent \textbf{Segment Length Bias Comparisons.}
We examine the segment length bias using the Jensen–Shannon Distance (JSD) metric proposed in HVQ~\cite{spurio2025hierarchical}. JSD measures how closely the predicted segment length distribution matches the ground truth distribution for each video by comparing their histograms (with 20-frame bins) using the Jensen–Shannon Distance. These distances are averaged per activity and then combined into a frame-weighted average across all activities to produce the final score. JSD quantifies the gap between the predicted and ground truth segment length distributions, where a smaller value indicates less bias in segment durations. From the results in Tab.~\ref{tab:jsd_comparison}, our HiST-VQ approach obtains the best overall results across all datasets, outperforming the baseline methods SMQ~\cite{gokay2025skeleton} and HVQ~\cite{spurio2025hierarchical}. The results confirm that our approach reduces segment length bias and better captures the variability of action segments. Fig.~\ref{fig:histogram} shows example histograms of segment lengths on BABEL Subset-3. SMQ~\cite{gokay2025skeleton} and HVQ~\cite{spurio2025hierarchical} fail to capture the shortest segments, as reflected in the first bin. Additionally, SMQ predicts excessive segments in the second bin. In contrast, our histogram distribution closely matches the ground truth.

\begin{table}[tb]
    \caption{Segment length bias comparisons. Lower is better.}
    \label{tab:jsd_comparison}
    \centering
    \setlength{\tabcolsep}{5pt}
    \resizebox{\columnwidth}{!}{
    \begin{tabular}{@{}l|cccccc@{}}
        \hline
        \textbf{Method} & \textbf{HuGaDB} & \textbf{LARa} & \textbf{BABEL S1} & \textbf{BABEL S2} & \textbf{BABEL S3} & \textbf{Avg.} \\
        \hline
        HVQ~\cite{spurio2025hierarchical} & 97.4 & 94.9 & 89.8 & 88.0 & 87.3 & 91.5 \\
        SMQ~\cite{gokay2025skeleton} & \textbf{87.1} & \underline{74.2} & \underline{72.8} & \underline{77.0} & \underline{81.9} & \underline{78.6} \\
        \cellcolor{beaublue}\textbf{HiST-VQ (Ours)} & \cellcolor{beaublue}\underline{89.0} & \cellcolor{beaublue}\textbf{73.8} & \cellcolor{beaublue}\textbf{72.4} & \cellcolor{beaublue}\textbf{71.6} & \cellcolor{beaublue}\textbf{80.9} & \cellcolor{beaublue}\textbf{77.5} \\
        \hline
    \end{tabular}
    }
\end{table}

\begin{figure}[tb]
    \centering
        \begin{subfigure}[b]{0.49\columnwidth}
            \includegraphics[trim= 40 0 85 0, clip, width=\linewidth]{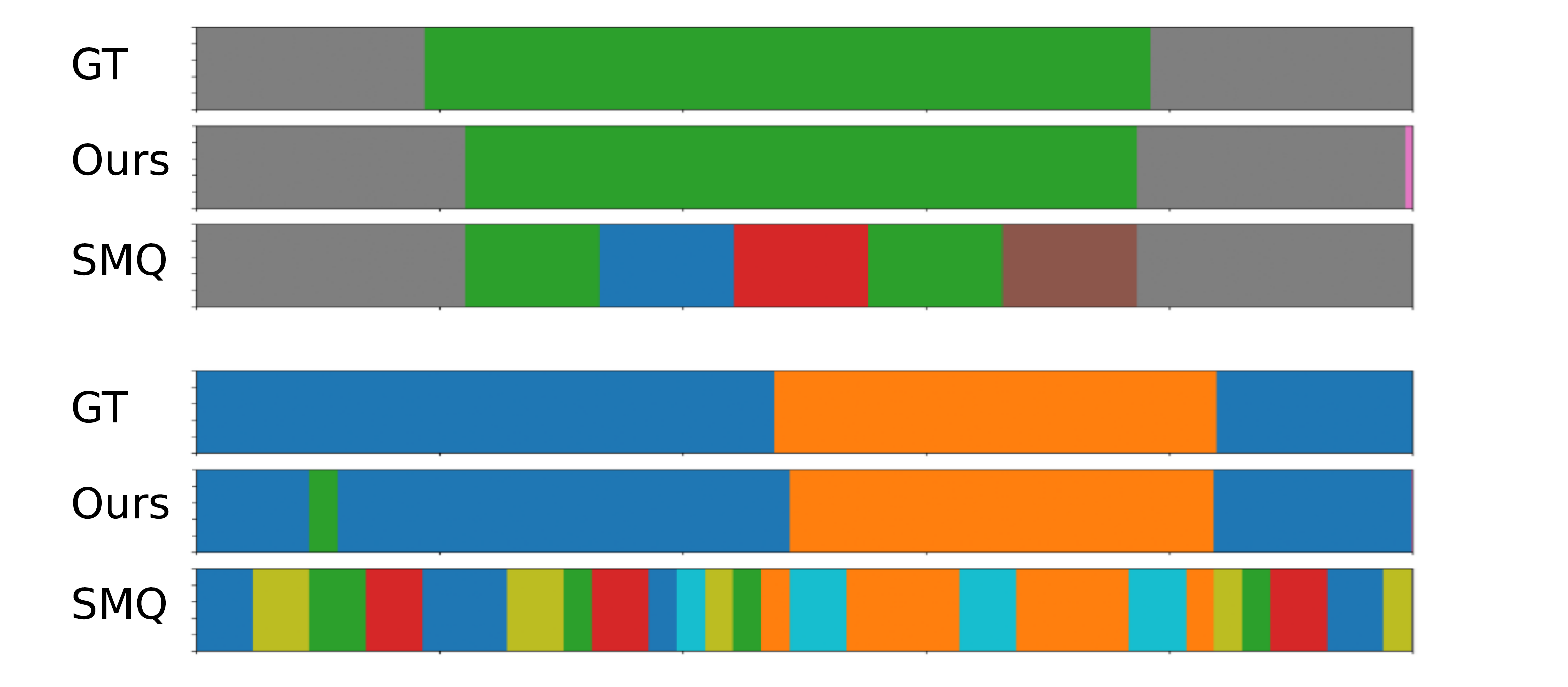}
            \caption{HuGaDB}
        \end{subfigure}\hfill
        \begin{subfigure}[b]{0.49\columnwidth}
            \includegraphics[trim= 40 0 85 0, clip, width=\linewidth]{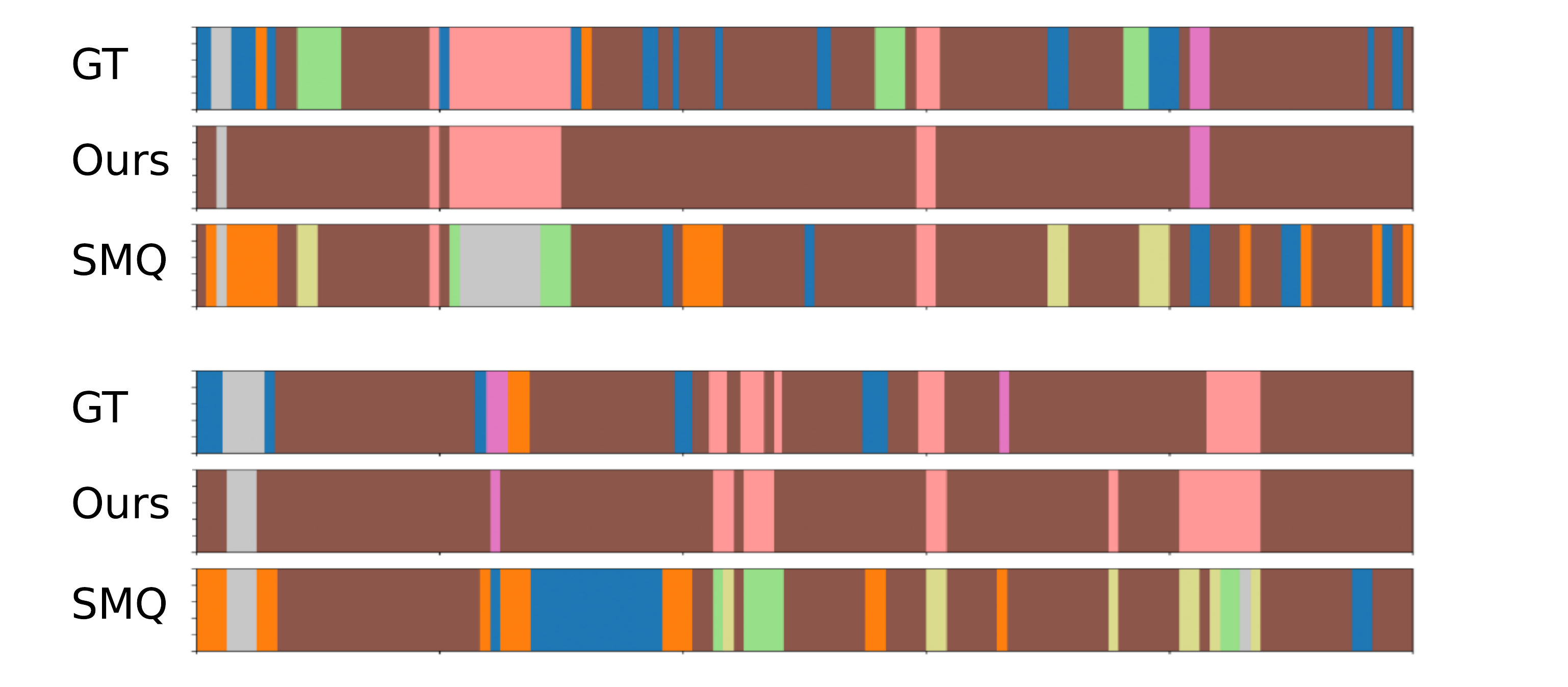}
            \caption{LARa}
        \end{subfigure}

        \begin{subfigure}[b]{0.49\columnwidth}
            \includegraphics[trim= 40 0 85 0, clip, width=\linewidth]{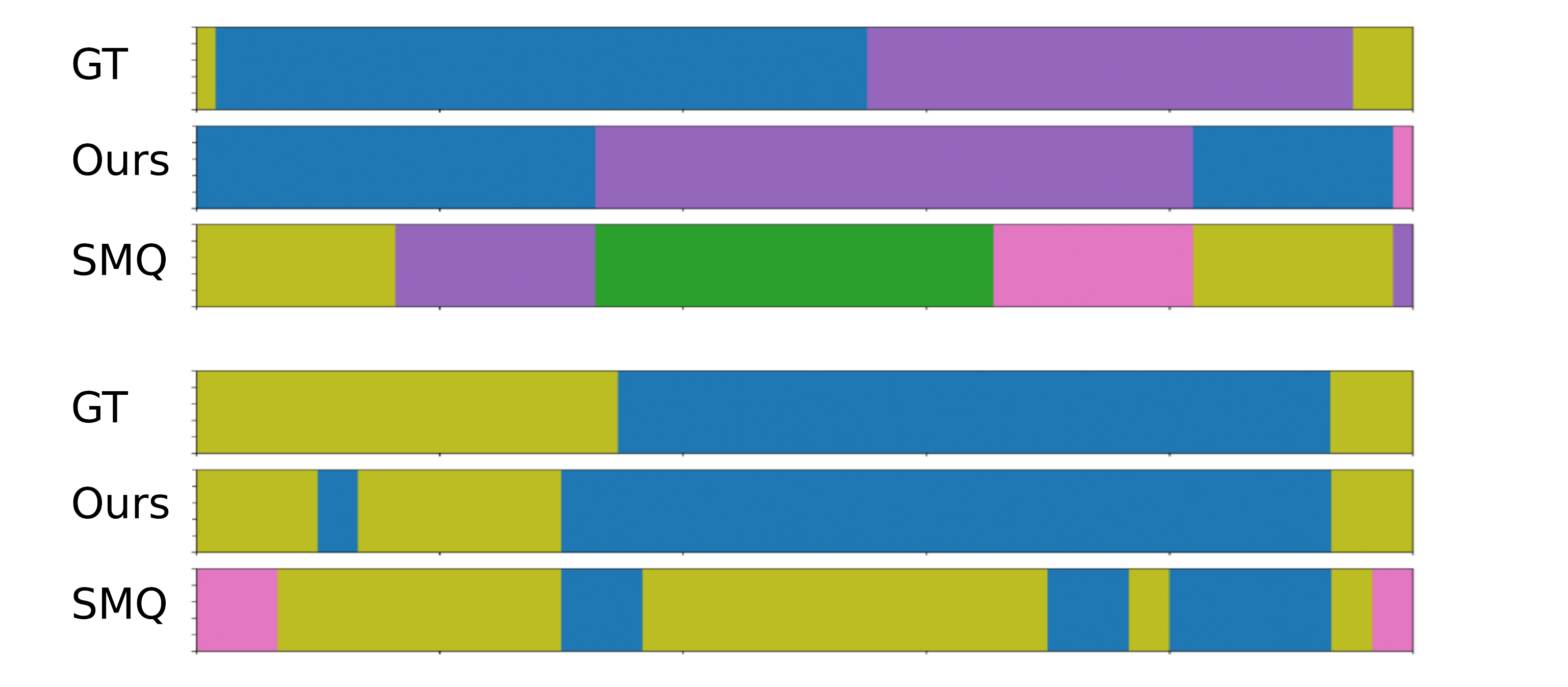}
            \caption{BABEL Subset-2}
        \end{subfigure}\hfill
        \begin{subfigure}[b]{0.49\columnwidth}
            \includegraphics[trim= 40 0 85 0, clip, width=\linewidth]{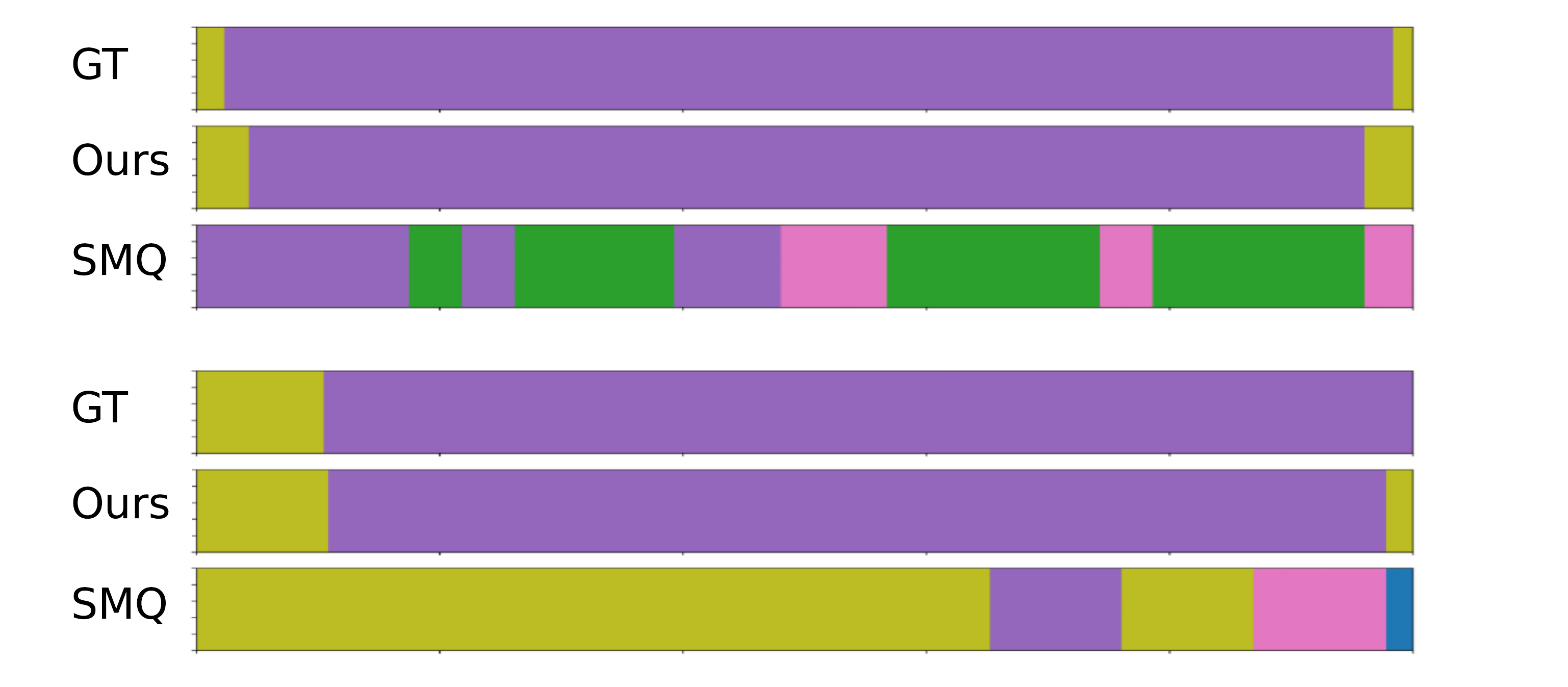}
            \caption{BABEL Subset-3}
        \end{subfigure}
    \caption{Qualitative comparisons.}
    \label{fig:qualitative}
\end{figure}

\noindent \textbf{Qualitative Comparisons.}
Fig.~\ref{fig:qualitative} plots example qualitative results of our HiST-VQ approach and the baseline method SMQ~\cite{gokay2025skeleton} on HuGaDB, LARa, BABEL Subset-2, and BABEL Subset-3. From the results, SMQ~\cite{gokay2025skeleton} has a tendency to over-segment the sequences, which is especially visible in the HuGaDB and LARa examples, where it predicts far more segments than present in the ground truth. Furthermore, its segments align less accurately with the ground truth. In contrast, the alignment and number of segments predicted by HiST-VQ are much closer to the ground truth across all examples. This highlights the superior accuracy and reduced segment length bias of our HiST-VQ approach compared to the state-of-the-art model SMQ~\cite{gokay2025skeleton}.

\subsection{Ablation Results}
\label{sec:ablations}

\begin{table}[tb]
    \caption{Impact of model components.}
    \label{tab:ablation_components}
    \centering
    \setlength{\tabcolsep}{5pt}
    \resizebox{\columnwidth}{!}{
    \begin{tabular}{@{}l|lllll|lllll@{}}
        \hline
        & \multicolumn{5}{c|}{\textbf{HuGaDB}} & \multicolumn{5}{c}{\textbf{BABEL Subset-3}} \\
        \cline{2-11}
        \textbf{Method} & \textbf{MoF} & \textbf{Edit} & \multicolumn{3}{c|}{\textbf{F1@\{10, 25, 50\}}} & \textbf{MoF} & \textbf{Edit} & \multicolumn{3}{c}{\textbf{F1@\{10, 25, 50\}}} \\
        \hline
        \cellcolor{beaublue}\textbf{All} & \cellcolor{beaublue}\textbf{48.2} & \cellcolor{beaublue}\textbf{44.3} & \cellcolor{beaublue}\textbf{49.4} & \cellcolor{beaublue}\textbf{39.8} & \cellcolor{beaublue}\underline{28.3} & \cellcolor{beaublue}\textbf{44.0} & \cellcolor{beaublue}\textbf{41.8} & \cellcolor{beaublue}\textbf{47.1} & \cellcolor{beaublue}\textbf{33.2} & \cellcolor{beaublue}\textbf{24.1} \\
        w/o Spat Recon & 28.5 & 26.7 & 26.5 & 17.8 & 6.9 & 36.7 & \underline{37.3} & \underline{41.6} & 24.6 & 15.5 \\
        w/o Commit & \underline{45.8} & 38.6 & 43.4 & 37.4 & \textbf{28.7} & \underline{43.3} & 29.0 & 37.8 & 22.7 & 15.3 \\
        w/o Temp Recon & 45.6 & \underline{43.1} & \underline{45.8} & \underline{38.2} & 26.5 & 42.8 & 30.3 & 40.4 & \underline{26.6} & \underline{20.2} \\
        
        \hline
    \end{tabular}
    }
\end{table}
\noindent \textbf{Impact of Model Components.}
We systematically remove one model component at a time to assess its contribution to the overall performance and report the results on HuGaDB and BABEL Subset-3 in Tab.~\ref{tab:ablation_components}. It is evident from the results that the top configuration which includes all components, i.e., commitment loss for hierarchical clustering, spatial reconstruction loss, and temporal reconstruction loss, produces the best overall performance. Next, removing the spatial reconstruction loss leads to the biggest drop in performance, which indicates its critical role in representation learning. Lastly, eliminating the temporal reconstruction or commitment loss reduces the performance notably, which confirms their contribution to the overall performance.

\begin{table}[tb]
    \caption{Impact of $\alpha$.}
    \label{tab:ablation_alpha}
    \centering
    \setlength{\tabcolsep}{5pt}
    \resizebox{\columnwidth}{!}{
    \begin{tabular}{@{}c|lllll|lllll@{}}
        \hline
        & \multicolumn{5}{c|}{\textbf{HuGaDB}} & \multicolumn{5}{c}{\textbf{BABEL Subset-3}} \\
        \cline{2-11}
        $\boldsymbol{\alpha}$ & \textbf{MoF} & \textbf{Edit} & \multicolumn{3}{c|}{\textbf{F1@\{10, 25, 50\}}} & \textbf{MoF} & \textbf{Edit} & \multicolumn{3}{c}{\textbf{F1@\{10, 25, 50\}}} \\
        \hline
        1 & \underline{47.1} & \underline{42.8} & \underline{47.9} & \textbf{40.0} & \textbf{28.6} & 42.4 & \underline{33.7} & \underline{41.8} & \underline{28.1} & \underline{20.4} \\
        \cellcolor{beaublue}\textbf{2} & \cellcolor{beaublue}\textbf{48.2} & \cellcolor{beaublue}\textbf{44.3} & \cellcolor{beaublue}\textbf{49.4} & \cellcolor{beaublue}\underline{39.8} & \cellcolor{beaublue}\underline{28.3} & \cellcolor{beaublue}\underline{44.0} & \cellcolor{beaublue}\textbf{41.8} & \cellcolor{beaublue}\textbf{47.1} & \cellcolor{beaublue}\textbf{33.2} & \cellcolor{beaublue}\textbf{24.1} \\
        3 & 44.8 & 37.0 & 41.9 & 35.5 & 27.2 & \textbf{45.8} & 29.2 & 35.0 & 23.8 & 19.6 \\
        
        \hline
    \end{tabular}
    }
\end{table}
\noindent \textbf{Impact of $\alpha$.}
\cref{tab:ablation_alpha} presents the effect of varying $\alpha$ (i.e., the ratio between the number of subactions and the number of actions, as illustrated in Fig.~\ref{fig:method}) on HuGaDB and BABEL Subset-3. From \cref{tab:ablation_alpha}, increasing $\alpha$ from 1 to 2 improves most metrics, indicating that using a moderate number of subactions helps capture fine-grained clusters and enhances the overall performance. Further increasing $\alpha$ to 3 degrades the performance due to noises/distractions caused by having too many subactions. Overall, the results indicate that $\alpha=2$ provides the best overall performance.

\begin{table}[tb]
    \caption{Impact of number of hierarchy levels.}
    \label{tab:ablation_hierarchy}
    \centering
    \setlength{\tabcolsep}{5pt}
    \resizebox{\columnwidth}{!}{
    \begin{tabular}{@{}c|lllll|lllll@{}}
        \hline
        & \multicolumn{5}{c|}{\textbf{HuGaDB}} & \multicolumn{5}{c}{\textbf{BABEL Subset-3}} \\
        \cline{2-11}
        \textbf{Levels} & \textbf{MoF} & \textbf{Edit} & \multicolumn{3}{c|}{\textbf{F1@\{10, 25, 50\}}} & \textbf{MoF} & \textbf{Edit} & \multicolumn{3}{c}{\textbf{F1@\{10, 25, 50\}}} \\
        \hline        
        1 & \underline{45.7} & \underline{38.4} & \underline{43.5} & 34.9 & \underline{27.0} & 43.6 & \underline{35.4} & \underline{44.7} & \underline{30.7} & \underline{23.0} \\
        \cellcolor{beaublue}\textbf{2} & \cellcolor{beaublue}\textbf{48.2} & \cellcolor{beaublue}\textbf{44.3} & \cellcolor{beaublue}\textbf{49.4} & \cellcolor{beaublue}\textbf{39.8} & \cellcolor{beaublue}\textbf{28.3} & \cellcolor{beaublue}\underline{44.0} & \cellcolor{beaublue}\textbf{41.8} & \cellcolor{beaublue}\textbf{47.1} & \cellcolor{beaublue}\textbf{33.2} & \cellcolor{beaublue}\textbf{24.1} \\
        3 & \underline{45.7} & 36.9 & 43.0 & \underline{35.0} & 24.4 & \textbf{44.6} & 33.2 & 43.3 & 29.5 & 22.8 \\
        
        \hline
    \end{tabular}
    }
\end{table}
\noindent \textbf{Impact of Number of Hierarchy Levels.}
We evaluate different variants of our HiST-VQ model, which have one, two, or three hierarchy levels respectively and include the results on HuGaDB and BABEL Subset-3 in \cref{tab:ablation_hierarchy}. It can be seen from the results that utilizing a two-level hierarchy (for hierarchical clustering) outperforms using a single-level hierarchy (for flat clustering) across all metrics. Moreover, adding a third level to the hierarchy significantly reduces the results, because of noises/distractions induced by having too fine-grained subactions. Finally, the results highlight that a two-level hierarchy offers the strongest performance.

\begin{table}[tb]
    \caption{Impact of $\lambda_{spat}$.}
    \label{tab:ablation_lambda_recon}
    \centering
    \setlength{\tabcolsep}{5pt}
    \resizebox{\columnwidth}{!}{
    \begin{tabular}{@{}c|lllll|lllll@{}}
        \hline
        & \multicolumn{5}{c|}{\textbf{HuGaDB}} & \multicolumn{5}{c}{\textbf{BABEL Subset-3}} \\
        \cline{2-11}
        $\boldsymbol{\lambda_{spat}}$ & \textbf{MoF} & \textbf{Edit} & \multicolumn{3}{c|}{\textbf{F1@\{10, 25, 50\}}} & \textbf{MoF} & \textbf{Edit} & \multicolumn{3}{c}{\textbf{F1@\{10, 25, 50\}}} \\
        \hline
        
        0.0005 
        & 46.6 & 37.4 & 41.5 & 31.2 & 22.4 
        & 43.4 & \underline{40.8} & \textbf{49.5} & \textbf{34.0} & \underline{23.7} \\
        
        \cellcolor{beaublue}\textbf{0.001} 
        & \cellcolor{beaublue}\textbf{48.2} & \cellcolor{beaublue}\textbf{44.3} & \cellcolor{beaublue}\textbf{49.4} & \cellcolor{beaublue}\underline{39.8} & \cellcolor{beaublue}28.3 
        & \cellcolor{beaublue}\textbf{44.1} & \cellcolor{beaublue}\textbf{41.8} & \cellcolor{beaublue}\underline{47.1} & \cellcolor{beaublue}\underline{33.2} & \cellcolor{beaublue}\textbf{24.1} \\
        
        0.002 
        & 44.9 & \underline{43.8} & \underline{49.3} & \textbf{40.3} & \textbf{30.0} 
        & \underline{44.0} & 29.7 & 40.9 & 27.2 & 21.0 \\
        
        0.005 
        & \underline{47.5} & 41.0 & 47.4 & 38.6 & \underline{28.6} 
        & \underline{44.0} & 30.7 & 41.7 & 28.4 & 22.2 \\
    
        \hline
    \end{tabular}
    }
\end{table}
\noindent \textbf{Impact of $\lambda_{spat}$.}
We now analyze the impact of the weight $\lambda_{spat}$ for the spatial reconstruction loss on the model performance. In particular, we vary $\lambda_{spat}$ in the range of $[0.0005, 0.005]$ to assess its influence on the model performance on HuGaDB and BABEL Subset-3 and present the results in~\cref{tab:ablation_lambda_recon}. It is evident from the results that $\lambda_{spat} = 0.001$ produces the best overall performance.

\begin{table}[tb]
    \caption{Impact of $\lambda_{temp}$.}
    \label{tab:ablation_lambda_temp}
    \centering
    \setlength{\tabcolsep}{5pt}
    \resizebox{\columnwidth}{!}{
    \begin{tabular}{@{}c|lllll|lllll@{}}
        \hline
        & \multicolumn{5}{c|}{\textbf{HuGaDB}} & \multicolumn{5}{c}{\textbf{BABEL Subset-3}} \\
        \cline{2-11}
        $\boldsymbol{\lambda_{temp}}$ & \textbf{MoF} & \textbf{Edit} & \multicolumn{3}{c|}{\textbf{F1@\{10, 25, 50\}}} & \textbf{MoF} & \textbf{Edit} & \multicolumn{3}{c}{\textbf{F1@\{10, 25, 50\}}} \\
        \hline

        \cellcolor{beaublue}\textbf{0.02}
        & \cellcolor{beaublue}\textbf{48.2} & \cellcolor{beaublue}\textbf{44.3} & \cellcolor{beaublue}\textbf{49.4} & \cellcolor{beaublue}\textbf{39.8} & \cellcolor{beaublue}\underline{28.3} 
        & \cellcolor{beaublue}44.0 & \cellcolor{beaublue}\textbf{41.8} & \cellcolor{beaublue}\textbf{47.1} & \cellcolor{beaublue}\textbf{33.2} & \cellcolor{beaublue}\textbf{24.1} \\
        
        0.2
        & 46.3 & 38.5 & 45.1 & 36.6 & 26.4 
        & \textbf{45.8} & \underline{37.8} & \underline{45.6} & \underline{31.4} & 22.1 \\
    
        2 
        & 39.6 & \underline{42.4} & 45.0 & 34.2 & 22.3 
        & 43.2 & 28.8 & 39.5 & 26.1 & 20.4 \\
    
        20 
        & \underline{47.8} & 41.1 & \underline{48.1} & \underline{39.4} & \textbf{31.3} 
        & \underline{44.4} & 32.5 & 43.5 & 30.4 & \underline{24.0} \\
    
        \hline
    \end{tabular}
    }
\end{table}
\noindent \textbf{Impact of $\lambda_{temp}$.}
We study the effect of the weight $\lambda_{temp}$ for the temporal reconstruction loss on the model performance and report the results on HuGaDB and BABEL Subset-3 in \cref{tab:ablation_lambda_temp}. We train our model with four different values of $\lambda_{temp}$, in the range $[0.02, 20]$ to evaluate how it affects the overall results. ~\cref{tab:ablation_lambda_temp} shows that the model performs best at $\lambda_{temp} = 0.02$ across both datasets, and the performance degrades with higher values.

\begin{table}[tb]
    \caption{Impact of input to spatial decoder.}
    \label{tab:ablation_spatial_input}
    \centering
    \setlength{\tabcolsep}{5pt}
    \resizebox{\columnwidth}{!}{
    \begin{tabular}{@{}c|lllll|lllll@{}}
        \hline
        
        & \multicolumn{5}{c|}{\textbf{HuGaDB}} & \multicolumn{5}{c}{\textbf{BABEL Subset-3}} \\
        \cline{2-11}
        \textbf{Input} & \textbf{MoF} & \textbf{Edit} & \multicolumn{3}{c|}{\textbf{F1@\{10, 25, 50\}}} & \textbf{MoF} & \textbf{Edit} & \multicolumn{3}{c}{\textbf{F1@\{10, 25, 50\}}} \\
        \hline
        $\mathbf{Q}^Z$ & 46.2 & 39.2 & \underline{46.5} & \underline{38.4} & \textbf{29.9} & \textbf{45.6} & 26.4 & 33.9 & 23.4 & 17.8 \\
        \cellcolor{beaublue}$\mathbf{Q}^A$ & \cellcolor{beaublue}\textbf{48.2} & \cellcolor{beaublue}\textbf{44.3} & \cellcolor{beaublue}\textbf{49.4} & \cellcolor{beaublue}\textbf{39.8} & \cellcolor{beaublue}\underline{28.3} & \cellcolor{beaublue}\underline{44.0} & \cellcolor{beaublue}\textbf{41.8} & \cellcolor{beaublue}\textbf{47.1} & \cellcolor{beaublue}\textbf{33.2} & \cellcolor{beaublue}\textbf{24.1} \\
        Both & \underline{46.6} & \underline{42.2} & 46.3 & 38.1 & \underline{28.3} & \underline{44.0} & \underline{31.0} & \underline{41.5} & \underline{28.5} & \underline{22.5} \\
        
        \hline
    \end{tabular}
    }
\end{table}
\noindent \textbf{Impact of Input to Spatial Decoder.}
To analyze the effect of the spatial decoder placement in the model architecture in Fig.~\ref{fig:method}, we pass different vector quantization outputs (i.e., $\mathbf{Q}^Z$, $\mathbf{Q}^A$, or both)  to the spatial decoder and include the results on HuGaDB and BABEL Subset-3 in Tab.~\ref{tab:ablation_spatial_input}. We observe from Tab.~\ref{tab:ablation_spatial_input} that the spatial decoder performs the best when the output of the second vector quantization $\mathbf{Q}^A$ is used as its input. Next, when the output of the first vector quantization $\mathbf{Q}^Z$ is provided instead, the performance drops across most metrics. Lastly, when both $\mathbf{Q}^Z$ and $\mathbf{Q}^A$  are used, we see a similar performance drop across all metrics.

\begin{table}[tb]
    \caption{Impact of input to temporal decoder.}
    \label{tab:ablation_temporal_input}
    \centering
    \setlength{\tabcolsep}{5pt}
    \resizebox{\columnwidth}{!}{
    \begin{tabular}{@{}c|lllll|lllll@{}}
        \hline
        & \multicolumn{5}{c|}{\textbf{HuGaDB}} & \multicolumn{5}{c}{\textbf{BABEL Subset-3}} \\
        \cline{2-11}
        \textbf{Input} & \textbf{MoF} & \textbf{Edit} & \multicolumn{3}{c|}{\textbf{F1@\{10, 25, 50\}}} & \textbf{MoF} & \textbf{Edit} & \multicolumn{3}{c}{\textbf{F1@\{10, 25, 50\}}} \\
        \hline        
        \cellcolor{beaublue}$\mathbf{Q}^Z$ & \cellcolor{beaublue}\textbf{48.2} & \cellcolor{beaublue}\textbf{44.3} & \cellcolor{beaublue}\textbf{49.4} & \cellcolor{beaublue}\textbf{39.8} & \cellcolor{beaublue}\underline{28.3} & \cellcolor{beaublue}\underline{44.0} & \cellcolor{beaublue}\textbf{41.8} & \cellcolor{beaublue}\textbf{47.1} & \cellcolor{beaublue}\textbf{33.2} & \cellcolor{beaublue}\textbf{24.1} \\
        
        $\mathbf{Q}^A$ & \underline{48.0} & 37.5 & 44.1 & 36.1 & 26.6 & \textbf{44.7} & 31.1 & \underline{42.1} & \underline{28.0} & \underline{20.9} \\
        
        Both & 47.0 & \underline{41.7} & \underline{46.7} & \underline{37.9} & \textbf{28.8} & 43.2 & \underline{32.9} & 41.2 & 26.5 & 18.2 \\
        
        \hline
    \end{tabular}
    }
\end{table}
\noindent \textbf{Impact of Input to Temporal Decoder.}
We examine the impact of feeding various vector quantization outputs (i.e., $\mathbf{Q}^Z$, $\mathbf{Q}^A$, or both) as input to the temporal decoder in Fig.~\ref{fig:method} and present the results on HuGaDB and BABEL Subset-3 in Tab.~\ref{tab:ablation_temporal_input}. As evident from Tab.~\ref{tab:ablation_temporal_input}, the best performance is achieved when the output of the first vector quantization $\mathbf{Q}^Z$ is used as input to the temporal decoder, outperforming using the second vector quantization $\mathbf{Q}^A$ or both $\mathbf{Q}^Z$ and $\mathbf{Q}^A$ as input to the temporal decoder. This is likely because temporal reconstruction is a simpler task than spatial reconstruction. Thus, low-level subaction representations are sufficient for temporal reconstruction, while high-level action representations are necessary for spatial reconstruction.

\noindent \textbf{Supplementary Material.} Extra implementation details and experiment results are in the supplementary material.
\section{Conclusion}
\label{sec:conclusion}

We present an unsupervised skeleton-based temporal action segmentation approach built upon a novel hierarchical spatiotemporal vector quantization. We first develop a hierarchical approach with a two-level vector quantization, i.e., skeletons are assigned to fine-grained subactions at the lower level, while subactions are mapped to actions at the higher level. Our hierarchical approach achieves better results than the non-hierarchical counterpart, while relying mostly on spatial cues via skeleton reconstruction. Moreover, we enhance our approach by incorporating both spatial and temporal cues, forming a hierarchical spatiotemporal vector quantization framework, i.e., multi-level clustering is conducted while both input skeletons and associated timestamps are recovered. Finally, we perform evaluations on HuGaDB, LARa, and BABEL to show our superior results and less segment length bias over prior works. Our future works will incorporate skeleton augmentation~\cite{kwon2022context,tran2024learning} and explore applications in action tokenization~\cite{chen2025moto,vuong2025action}.

\noindent \textbf{LLM Usage Statement}
No large language models (LLMs) or generative AI tools were used in the preparation of this manuscript, including its writing, methodology, experiments, or analysis.

\appendix
\section{Supplementary Material}

In this supplementary material, we first provide the implementation details of our HiST-VQ model in Sec.~\ref{sec:supp_implementation_details}. Next, we present additional qualitative results across all datasets, i.e., HuGaDB~\cite{chereshnev2017hugadb}, LARa~\cite{niemann2020lara}, BABEL~\cite{punnakkal2021babel} Subset-1, Subset-2, and Subset-3 in Sec.~\ref{sec:supp_qualitative_results}. We then report results on the PKU-MMD v2 dataset and comparisons with other models such as CTE~\cite{kukleva2019unsupervised}, TOT~\cite{kumar2022unsupervised}, ASOT~\cite{xu2024temporally}, and SMQ~\cite{gokay2025skeleton} in Sec.~\ref{sec:supp_pkummd_results}, followed by sequence-level action segmentation results in Sec.~\ref{sec:supp_per_sequence_results}. Run time comparisons between HiST-VQ and competing methods are presented in Sec.~\ref{sec:supp_runtime_comparison}. Then we analyze the impact of the number of actions ($K$) on the model performance in Sec.~\ref{sec:supp_k_analysis}. Finally, we evaluate the robustness of our model to missing joints in Sec.~\ref{sec:supp_missing_joints}.

\subsection{Implementation Details}
\label{sec:supp_implementation_details}
\begin{table*}[t]
    \centering
    \caption{Hyperparameter settings for our HiST-VQ approach. Note that H, L, B denote HuGaDB, LARa, and BABEL respectively.}
    \setlength{\tabcolsep}{8pt}
    \begin{tabular}{@{}lr@{}}
        \toprule
        Hyperparameter & Value \\
        \midrule
        Batch size ($N$) & $8$ (H, L), $32$ (B)\\
        Number of joints ($V$) & $6$ (H), $22$ (L), $25$ (B)\\
        Joint dimension ($C$) & $6$ (H, L), $3$ (B)\\
        Patch size ($P$) & $60$ (H), $50$ (L), $30$ (B)\\
        Number of actions ($K$) & $10$ (H), $8$ (L), $5$ (B)\\
        Latent dimension ($D$) & $16$\\
        Hierarchy ratio ($\alpha$) & $2$\\
        EMA decay weight ($\beta$) & $0.5$\\
        Codebook usage threshold ($\nu_z$) & $3$\\
        Codebook usage threshold ($\nu_a$) & $1$\\
        Commitment loss weight ($\lambda_{commit}$) & $1.0$\\
        Spatial reconstruction loss weight ($\lambda_{spat}$) & $0.001$ \\
        Temporal reconstruction loss weight ($\lambda_{temp}$) & $0.02$ \\
        ADAM learning rate & $5\times10^{-4}$\\
        \bottomrule
    \end{tabular}
    \label{tab:hyperparameters}
\end{table*}
We implement our encoder and spatial decoder based on MS-TCN~\cite{farha2019ms}. In particular, we use a two-stage temporal convolutional network with three dilated residual layers per stage and 128 feature maps as our encoder and spatial decoder. The encoder produces joint-disentangled latent representations with the latent dimension $D = 16$. We convert skeleton-wise sequences to patch-wise representations with the patch size $P = 60$ frames for HuGaDB, $P = 50$ frames for LARa, and $P = 30$ frames for BABEL, corresponding to 1 second for each dataset, following SMQ~\cite{gokay2025skeleton}. We then feed the patches to our two-level vector quantization module with $2K$ subaction prototypes in the first level and $K$ action prototypes in the second level. $K$ is the ground truth number of actions. The codebooks are updated via EMA~\cite{van2017neural} with a decay weight $\beta = 0.5$. We set codebook usage thresholds $\nu_z = 3$ and $\nu_a = 1$. Quantized action patches $\mathbf{Q}^A$ and quantized subaction patches $\mathbf{Q}^Z$ are passed to our spatial and temporal decoders respectively. We employ an MLP network with two hidden layers (hidden dimensions $2 \times D \times V$ and $D \times V$ respectively) as our temporal decoder. We set $\lambda_{commit} = 1$, $\lambda_{spat} = 0.001$ and $\lambda_{temp} = 0.02$ for all datasets.. We follow SMQ~\cite{gokay2025skeleton} to set the batch size $N = 8$ for HuGaDB and LARa, and $N = 32$ for BABEL. Our model is implemented in PyTorch and is trained using ADAM~\cite{kingma2014adam} with a learning rate of $5\times10^{-4}$. Our experiments are conducted on a single NVIDIA RTX 3090 Ti (24GB). Our hyperparameter settings are summarized in Tab.~\ref{tab:hyperparameters}.

\begin{figure}[tb]
    \centering
        \begin{subfigure}[b]{0.49\columnwidth}
            \includegraphics[trim= 40 0 85 0, clip, width=\columnwidth]{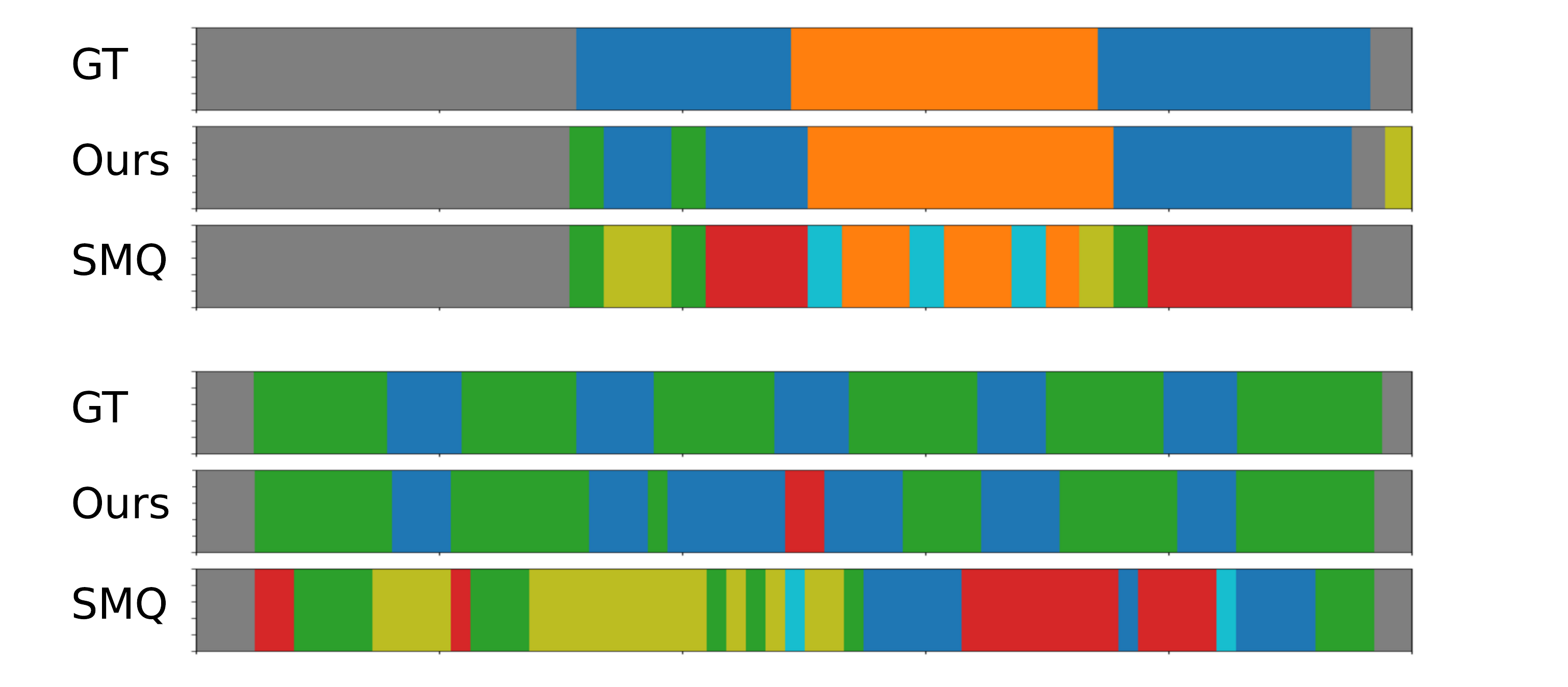}
            \caption{HuGaDB}
            \label{fig:supp_qualitative_hugadb}
        \end{subfigure}\hfill
        \begin{subfigure}[b]{0.49\columnwidth}
            \includegraphics[trim= 40 0 85 0, clip, width=\columnwidth]{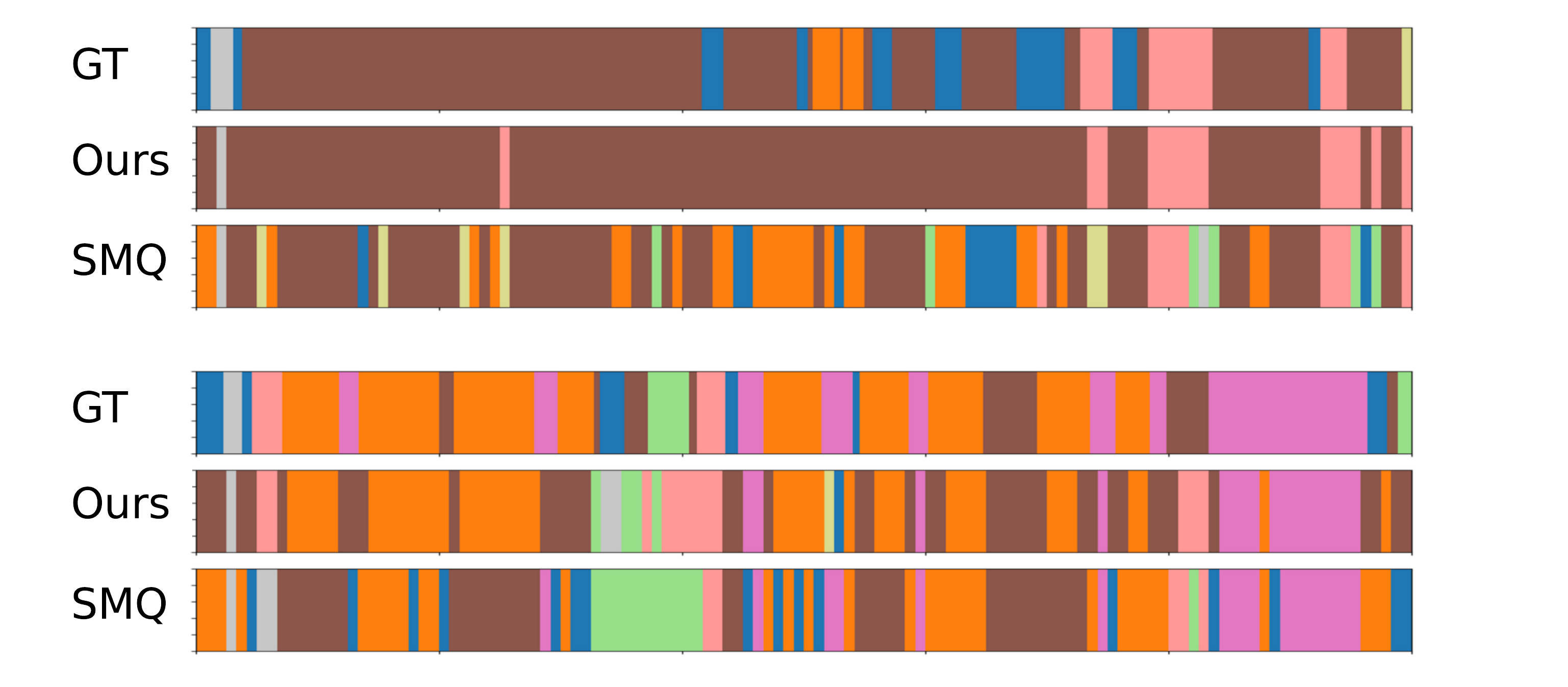}
            \caption{LARa}
            \label{fig:supp_qualitative_lara}
        \end{subfigure}

        \begin{subfigure}[b]{0.49\columnwidth}
            \includegraphics[trim= 40 0 85 0, clip, width=\columnwidth]{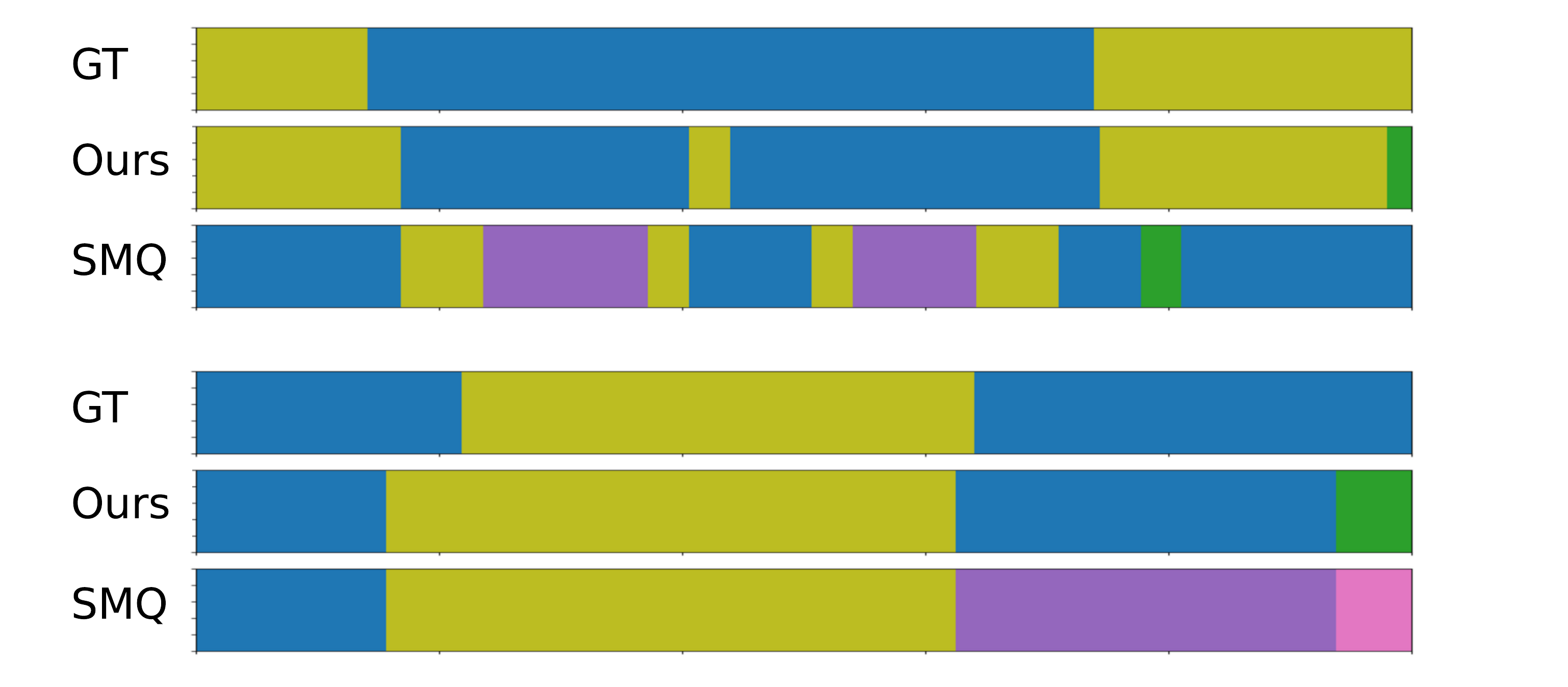}
            \caption{BABEL Subset-1}
        \end{subfigure}\hfill
        \begin{subfigure}[b]{0.49\columnwidth}
            \includegraphics[trim= 40 0 85 0, clip, width=\columnwidth]{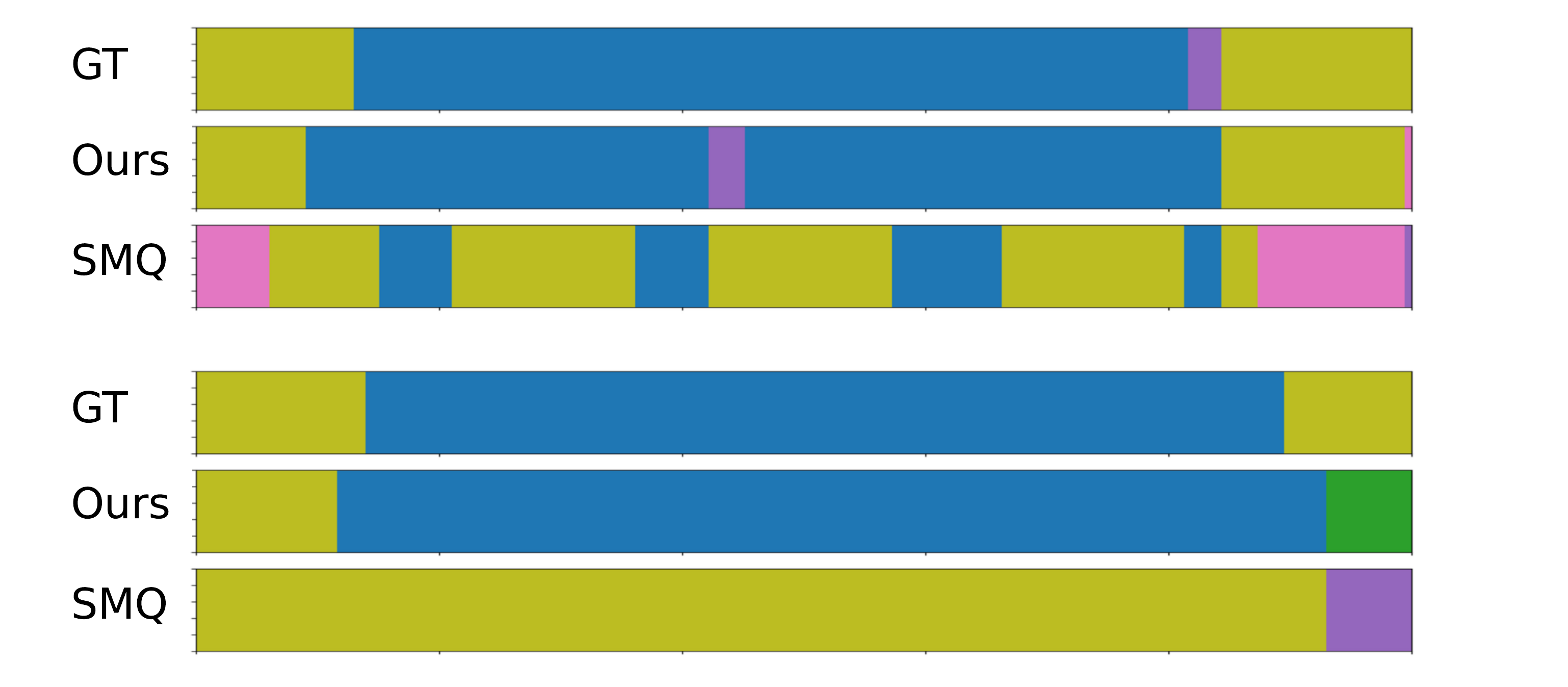}
            \caption{BABEL Subset-2}
        \end{subfigure}

        \begin{subfigure}[b]{0.49\columnwidth}
            \centering
            \includegraphics[trim= 40 0 85 0, clip, width=\columnwidth]{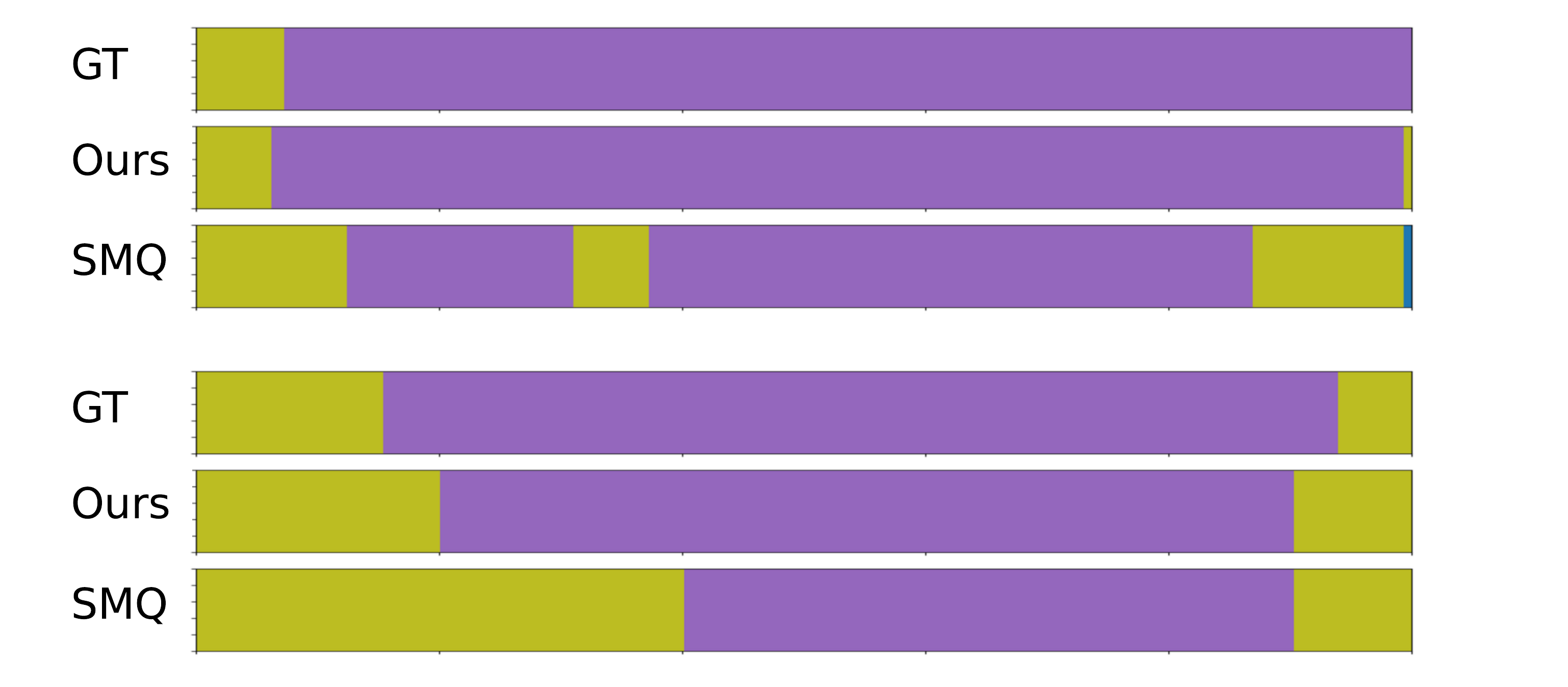}
            \caption{BABEL Subset-3}
        \end{subfigure}
    \caption{Qualitative comparisons.}
    \label{fig:supp_qualitative}
\end{figure}
\subsection{Qualitative Results}
\label{sec:supp_qualitative_results}
\cref{fig:supp_qualitative} presents additional qualitative results on all datasets, including BABEL Subset-1, which was not included in Fig.~4 of the main paper. As shown in \cref{fig:supp_qualitative}, the baseline method SMQ~\cite{gokay2025skeleton} frequently produces a large number of short segments, exhibiting over-segmentation. This often results in predicted action boundaries that are misaligned with the ground truth transitions between actions, particularly evident in \cref{fig:supp_qualitative_hugadb,fig:supp_qualitative_lara}. In contrast, our HiST-VQ model produces predictions that follow the ground truth segmentations more closely. The number of predicted segments is more consistent with the ground truth, and the segment boundaries better match the actual action transitions. These additional qualitative examples further demonstrate the robustness of HiST-VQ for action segmentation across all datasets.

\subsection{PKU-MMD v2 Results}
\label{sec:supp_pkummd_results}
\begin{table}[tb]
    \caption{Results on PKU-MMD v2.}
    \label{tab:pkummd_comparison}
    \centering
    \footnotesize
    \setlength{\tabcolsep}{5pt}
    \begin{tabular}{@{}l|ccccc@{}}
        \hline
        \textbf{Method} & \textbf{MoF} & \textbf{Edit} & \multicolumn{3}{c}{\textbf{F1@\{10, 25, 50\}}} \\
        \hline
        CTE~\cite{kukleva2019unsupervised} & 8.6 & 4.5 & 1.8 & 1.0 & 0.4 \\
        CTE+Viterbi~\cite{kukleva2019unsupervised} & 8.1 & 10.8 & 3.4 & 2.3 & 1.0 \\
        TOT~\cite{kumar2022unsupervised} & 6.6 & 3.0 & 0.6 & 0.1 & 0.0 \\
        TOT+Viterbi~\cite{kumar2022unsupervised} & \underline{15.1} & 10.8 & 5.8 & 4.2 & 2.2 \\
        ASOT~\cite{xu2024temporally} & 9.0 & 9.4 & 6.0 & 4.4 & 2.4 \\
        SMQ~\cite{gokay2025skeleton} & 13.2 & \underline{13.8} & \underline{13.8} & \underline{10.6} & \textbf{5.6} \\
        \cellcolor{beaublue}\textbf{HiST-VQ (Ours)} & \cellcolor{beaublue}\textbf{22.0} & \cellcolor{beaublue}\textbf{22.6} & \cellcolor{beaublue}\textbf{16.7} & \cellcolor{beaublue}\textbf{11.5} & \cellcolor{beaublue}\underline{4.3} \\
        \hline
    \end{tabular}
\end{table}
We follow SMQ~\cite{gokay2025skeleton} to further evaluate the action segmentation capabilities of our HiST-VQ approach on the PKU-MMD v2 dataset. This dataset is especially challenging as it contains 41 action categories and a varying number of subjects per sequence. It is comprised of 1009 skeleton sequences lasting 1-2 minutes each at 30 FPS. Each sequence has from 1 to 2 subjects with 25 joints per subject. We compare HiST-VQ with other unsupervised action segmentation methods in \cref{tab:pkummd_comparison}. It is evident from the results that our model shows (substantial) improvement over the previous state-of-the-art, having gained 6.9\% over TOT+Viterbi~\cite{kumar2022unsupervised} on the MOF metric and 8.8\% over SMQ~\cite{gokay2025skeleton} on the Edit metric. It also outperforms SMQ~\cite{gokay2025skeleton} on the F1@10 and F1@25 scores by 2.9\% and 0.9\% respectively. The above results further confirm the advantage of our hierarchical clustering and spatiotemporal reconstruction modules over previous methods, even on datasets with multiple subjects per sequence and a large variety of actions.

\subsection{Per-Sequence Results}
\label{sec:supp_per_sequence_results}
\begin{table*}[tb]
    \caption{Results of sequence-level skeleton-based action segmentation.}
    \label{tab:Supp_per_sequence}
    \centering
    \footnotesize
    \setlength{\tabcolsep}{5pt}
    \begin{tabular}{@{}l|lllll|lllll@{}}
        \hline
        & \multicolumn{5}{c|}{\textbf{HuGaDB}} & \multicolumn{5}{c}{\textbf{LARa}} \\
        \cline{2-11}
        \textbf{Method} & \textbf{MoF} & \textbf{Edit} & \multicolumn{3}{c|}{\textbf{F1@\{10, 25, 50\}}} & \textbf{MoF} & \textbf{Edit} & \multicolumn{3}{c}{\textbf{F1@\{10, 25, 50\}}} \\
        \hline
        \hspace{1em}FINCH~\cite{sarfraz2019efficient} & 54.2 & 19.5 & 5.9 & 3.6 & 3.1 & 43.7 & 38.0 & 36.2 & 29.1 & 17.6 \\
        \hspace{1em}TW-FINCH~\cite{sarfraz2021temporally} & 57.5 & 39.8 & 45.8 & 36.4 & 25.8 & 37.5 & 17.8 & 22.7 & 12.2 & 4.0 \\
        \hspace{1em}TSA (K-means)~\cite{bueno2023leveraging} & 58.0 & 22.8 & 10.6 & 7.3 & 5.2 & 36.6 & 18.5 & 22.1 & 12.3 & 4.8 \\
        \hspace{1em}TSA (FINCH)\cite{bueno2023leveraging} & \underline{61.6} & 37.3 & 17.1 & 13.8 & 10.2 & 39.8 & 16.3 & 22.0 & 12.7 & 5.2 \\
        \hspace{1em}TSA (TW-FINCH)~\cite{bueno2023leveraging} & 54.1 & 38.5 & 37.4 & 27.4 & 17.6 & 36.4 & 14.1 & 22.6 & 11.8 & 4.2 \\
        \hspace{1em}SMQ~\cite{gokay2025skeleton} & 55.3 & \underline{44.5} & \underline{52.8} & \underline{44.6} & \underline{31.7} & \underline{51.7} & \underline{45.1} & \underline{45.3} & \underline{38.8} & \textbf{23.6} \\
        \cellcolor{beaublue}\hspace{1em}\textbf{HiST-VQ (Ours)} & \cellcolor{beaublue}\textbf{63.4} & \cellcolor{beaublue}\textbf{51.9} & \cellcolor{beaublue}\textbf{59.3} & \cellcolor{beaublue}\textbf{52.6} & \cellcolor{beaublue}\textbf{37.1} & \cellcolor{beaublue}\textbf{55.3} & \cellcolor{beaublue}\textbf{47.6} & \cellcolor{beaublue}\textbf{46.7} & \cellcolor{beaublue}\textbf{39.6} & \cellcolor{beaublue}\underline{23.0} \\
        \hline
    \end{tabular}
\end{table*}
We additionally evaluate the sequence-level action segmentation performance of our HiST-VQ model. Following the common protocol defined in prior works~\cite{gokay2025skeleton, bueno2023leveraging, sarfraz2019efficient, sarfraz2021temporally}, we perform local Hungarian matching for each sequence to determine the mapping between the predicted clusters and ground truth clusters, and provide the average number of unique actions per sequence computed across the dataset as $K$. As presented in \cref{tab:Supp_per_sequence}, HiST-VQ achieves the best performance among all evaluated models across all metrics on HuGaDB. On LARa, it achieves the best performance across most metrics, with SMQ~\cite{gokay2025skeleton} performing slightly better in F1@50. These results demonstrate that our hierarchical spatiotemporal vector quantization framework HiST-VQ yield more accurate segmentation boundaries at the sequence level. Importantly, similar to SMQ~\cite{gokay2025skeleton}, our method only requires the average number of actions in the dataset rather than the number of actions per sequence, like the other sequence-level unsupervised action segmentation models that were evaluated.

\subsection{Run Time Comparisons}
\label{sec:supp_runtime_comparison}
\begin{table}[tb]
\caption{Run time comparisons (in minutes).}
\label{tab:Supp_runtime}
\centering
\small
\setlength{\tabcolsep}{3.45pt}
\begin{tabular}{l|cccc}
\hline
& \textbf{CTE~\cite{kukleva2019unsupervised}} & \textbf{HVQ~\cite{spurio2025hierarchical}} & \textbf{SMQ~\cite{gokay2025skeleton}} & \textbf{HiST-VQ (Ours) } \\
\hline
HuGaDB & 10.3 & 9.2 & 13.0 & 13.2 \\
LARa   & 50.9 & 38.8 & 57.9 & 58.1 \\
\hline
\end{tabular}
\end{table}
\cref{tab:Supp_runtime} compares the run time of several unsupervised skeleton-based action segmentation methods, including CTE~\cite{kukleva2019unsupervised}, HVQ~\cite{spurio2025hierarchical}, SMQ~\cite{gokay2025skeleton}, and our HiST-VQ approach, when processing the HuGaDB and LARa datasets on a single NVIDIA RTX 3090 Ti (24GB). It can be seen from \cref{tab:Supp_runtime} that HiST-VQ achieves improved segmentation results while maintaining run time performance comparable to SMQ~\cite{gokay2025skeleton}, the previous best method, demonstrating a strong balance between computational efficiency and segmentation accuracy.

\subsection{Impact of Number of Actions}
\label{sec:supp_k_analysis}
\begin{table}[tb]
    \caption{Effect of varying the number of actions $K$ on segmentation performance on LARa. The ground truth number of actions ($K=8$) is highlighted.}
    \label{tab:Supp_k_number_of_actions}
    \centering
    \setlength{\tabcolsep}{5pt}
    \resizebox{\columnwidth}{!}{
    \begin{tabular}{@{}l|ccccc@{}}
        \hline
        \textbf{Number of Actions ($K$)} & \textbf{MoF} & \textbf{Edit} & \multicolumn{3}{c}{\textbf{F1@\{10, 25, 50\}}} \\
        \hline
        3 & 44.92 & 34.08 & 34.81 & 27.23 & 15.14 \\
        4 & 40.74 & 36.93 & 34.34 & 27.46 & 15.04 \\
        5 & 42.41 & 34.79 & 35.31 & 28.92 & 16.35 \\
        6 & \underline{44.95} & 40.87 & \textbf{41.57} & \underline{33.86} & \underline{19.28} \\
        7 & 40.16 & \underline{43.80} & 40.14 & 32.64 & 18.45 \\
        \cellcolor{beaublue}8 (ground truth) & \cellcolor{beaublue}\textbf{45.90} & \cellcolor{beaublue}42.13 & \cellcolor{beaublue}\underline{41.06} & \cellcolor{beaublue}\textbf{33.99} & \cellcolor{beaublue}\underline{\textbf{19.34}} \\
        9 & 36.61 & \textbf{43.95} & 37.43 & 29.97 & 17.13 \\
        10 & 35.77 & 38.59 & 35.30 & 28.44 & 15.11 \\
        \hline
    \end{tabular}
    }
\end{table}
Unsupervised action segmentation methods are typically evaluated by providing the number of action classes $K$ present in a particular dataset. To examine the sensitivity of our HiST-VQ approach to this parameter, we vary the number of clusters and evaluate the model performance on the LARa dataset. The results are reported in \cref{tab:Supp_k_number_of_actions}. We observe that HiST-VQ demonstrates relatively stable performance across a wide range of $K$ values. Although the model achieves the best overall performance when the ground truth number of actions ($K=8$) is provided, it still produces competitive results even when $K$ is under or overestimated. Performance improves gradually as $K$ approaches the ground truth value and remains relatively stable above and below this value.

\subsection{Robustness to Missing Joints}
\label{sec:supp_missing_joints}
\begin{table}[tb]
    \caption{Robustness to missing joints on LARa.}
    \label{tab:ablation_missing_joints}
    \centering
    \setlength{\tabcolsep}{5pt}
    \resizebox{\columnwidth}{!}{
    \begin{tabular}{@{}l|lllll@{}}
        \hline
        \textbf{Setting} & \textbf{MoF} & \textbf{Edit} & \multicolumn{3}{c}{\textbf{F1@\{10, 25, 50\}}} \\
        \hline

        25\% joints missing (17 joints)
        & \underline{41.9} & \underline{40.3} & \underline{36.8} & \underline{29.4} & \underline{17.2} \\

        50\% joints missing (11 joints)
        & 35.6 & 39.5 & 34.8 & 28.0 & 16.0 \\

        Hand and wrist missing (18 joints)
        & 34.7 & 38.2 & 35.1 & 27.4 & 15.6 \\

        \cellcolor{beaublue}All joints (22 joints)
        & \cellcolor{beaublue}\textbf{45.9} & \cellcolor{beaublue}\textbf{42.1} & \cellcolor{beaublue}\textbf{41.1} & \cellcolor{beaublue}\textbf{34.0} & \cellcolor{beaublue}\textbf{19.3} \\

        \hline
    \end{tabular}
    }
\end{table}

We examine robustness to incomplete skeleton data on LARa by randomly dropping 25\% or 50\% of the joints per skeleton, and by removing the hand and wrist joints, following the settings of~\cite{gokay2025skeleton}. As shown in Tab.~\ref{tab:ablation_missing_joints}, HiST-VQ maintains relatively steady performance across all settings, with Edit staying within 4 points of the full-skeleton result. HiST-VQ also outperforms SMQ~\cite{gokay2025skeleton} on every metric under every missing-joint setting, e.g., by 16.4\% in MoF with 25\% of the joints missing. Notably, even with 25\% of the joints missing, HiST-VQ surpasses SMQ evaluated on the full skeleton on all metrics. These results demonstrate that HiST-VQ is robust to incomplete skeleton input.

{
\small
\bibliographystyle{ieeenat_fullname}
\bibliography{main}
}


\end{document}